\RequirePackage{fix-cm}
\documentclass[twocolumn]{svjour3}          % twocolumn
\usepackage{graphicx}
\emergencystretch=\maxdimen
\usepackage{natbib} % for citation
\usepackage{comment}
\usepackage{nicefrac}       % compact symbols for 1/2, etc.
\usepackage{microtype}      % microtypography

\usepackage{xcolor}
\usepackage{color, colortbl}
\usepackage{url}       % url citation
\usepackage{soul} % delete line
\usepackage{multirow}
\usepackage{graphicx}

\usepackage{caption}
\usepackage{subfigure}
\usepackage{array}
\newcolumntype{x}[1]{>{\centering\arraybackslash\hspace{0pt}}p{#1}}

\usepackage{fancyhdr}%导入fancyhdr包
\usepackage{amsmath}
\usepackage{amsfonts}
\usepackage{amssymb}
\usepackage{bbding}
\usepackage{pifont}
\usepackage[switch]{lineno} 

\newcommand{\myparagraph}[1]{\vspace{2pt}\noindent{\bf #1}}

\usepackage{balance}
\usepackage{epsfig}
\usepackage{graphicx}
\usepackage{amsmath}
\usepackage[american]{babel}
\usepackage{microtype}
\usepackage{xcolor}
\usepackage{color, colortbl}
\usepackage{multirow}

\usepackage{bm}
\usepackage{amsmath}
\usepackage{chngcntr}

\usepackage{paralist}
\usepackage[font=small]{caption}
 
\definecolor{cgreen}{RGB}{11,135,28}
\usepackage{algorithm,algpseudocode}
\usepackage{amsmath}
\usepackage{amsfonts}
 
\newcommand{\ie}{i.e.}
\newcommand{\eg}{e.g.}
\newcommand{\real}{\mathbb{R}}

\newcommand{\jimmyadd}[1]{\textcolor{black}{ #1}} % blue
\newcommand{\abcadd}[1]{\textcolor{black}{#1}} % red
\newcommand{\abcaddd}[1]{\textcolor{black}{#1}} % magenta
\newcommand{\abcaddpre}[1]{\textcolor{black}{#1}}

\newcommand{\wenjiaadd}[1]{\textcolor{black}{ #1}} 

\newcommand{\revise}[1]{\textcolor{black}{#1}}
\newcommand{\CUT}[1]{}
\begin{document}
% \linenumbers removed

\title{Group-based Distinctive Image Captioning with Memory Difference Encoding and Attention%\thanks{Grants or other notes
%about the article that should go on the front page should be
%placed here. General acknowledgments should be placed at the end of the article.}
}
% \subtitle{Do you have a subtitle?\\ If so, write it here}

%\titlerunning{Short form of title}        % if too long for running head

\author{Jiuniu~Wang \and Wenjia~Xu \and Qingzhong~Wang \and Antoni~B.~Chan
}

%\authorrunning{Short form of author list} % if too long for running head

\institute{
A.~B.~Chan is the corresponding author, and with the Department of Computer Science, City University of Hong Kong, Hong Kong.\\
J.~Wang is with Alibaba Group, Beijing, China, and also with Department of Computer Science, City University of Hong Kong, Hong Kong. \\ 
W.~Xu is with University of Chinese Academy of Sciences, Beijing, China. \\
Q.~Wang is with Department of Computer Science, City University of Hong Kong, Hong Kong. \\
\email{wangjiuniu.wjn@alibaba-inc.com, xuwenjia16@mails.ucas.ac.cn, qingzwang@outlook.com,  abchan@cityu.edu.hk} \\
% $^1$ The Department of Computer Science, City University of Hong Kong, Hong Kong\\
% $^{2}$ Alibaba Group, Beijing, China \\
%     $^{3}$ Aerospace Information Research Institute, Chinese Academy of Sciences, Beijing, China\\
%     $^{4}$ Baidu Research, Baidu Inc., Beijing, China\\
}

\date{Received: date / Accepted: date}
% The correct dates will be entered by the editor

\maketitle

\begin{abstract}
\revise{Recent advances in image captioning have focused on enhancing accuracy by substantially increasing the dataset and model size. While conventional captioning models exhibit high performance on established metrics such as BLEU, CIDEr, and SPICE, the capability of captions to distinguish the target image from other similar images is under-explored. To generate distinctive captions, a few pioneers employed contrastive learning or re-weighted the ground-truth captions. However, this approach often overlooks the relationships between objects in a similar image group~(\eg, items or properties within the same album or fine-grained events). In this paper, we introduce an innovative approach to enhance the distinctiveness of image captions, namely Group-based Differential Distinctive Captioning Method~(DifDisCap), which visually compares each image with other images in one similar group and highlights the uniqueness of each image. 
In particular, we introduce a Group-based Differential Memory Attention~(GDMA) module, designed to identify and emphasize object features that are uniquely distinguishable within an image group, i.e., those exhibiting low similarity with objects in other images.
This mechanism ensures that such unique object features are prioritized during caption generation, thereby enhancing the distinctiveness of the resulting captions.
To further refine this process, we select distinctive words from the ground-truth captions to guide both the language decoder and the GDMA module. Additionally, we propose a new evaluation metric, the Distinctive Word Rate (DisWordRate), to quantitatively assess caption distinctiveness. Quantitative results indicate that the proposed method significantly improves the distinctiveness of several baseline models, and achieves state-of-the-art performance on distinctiveness while not excessively sacrificing accuracy. Moreover, the results of user study agree with the quantitative evaluation and demonstrate the rationality of the new metric DisWordRate.}

\end{abstract}
\section{Introduction}

\label{intro}
\begin{figure}[tb]
	\begin{center}
		\includegraphics[width=\linewidth]{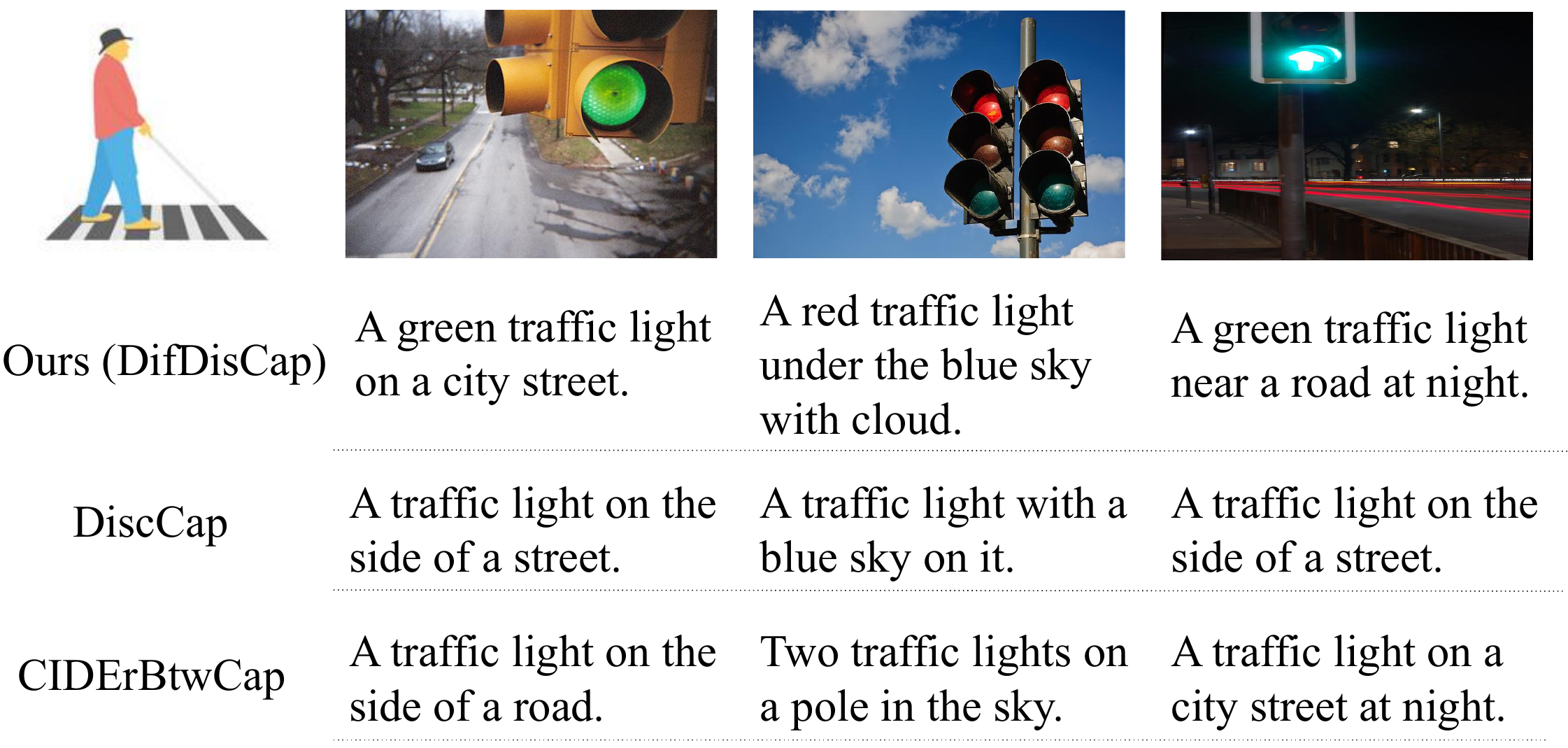}
	\end{center}
	\caption{\revise{Our model generates distinctive captions that can distinguish the target image from other similar images. Compared to current distinctive image captioning models such as DiscCap~\citep{2_disccap} and CIDErBtwCap~\citep{17_wang2020compare}, our captions can specify the important details, e.g., the color and the context of the traffic light, which can help a visually-impaired person to cross the street.}
    }
	%\abc{add citations for DisCap and CIDErBtwCap in the figure.}}
	\label{fig:teaser_figure}
\end{figure}

Acquiring knowledge through multiple sensory modalities, such as vision and language, is gaining considerable interest and facilitating the development of multimodal applications. Among them, the task of image captioning has drawn much attention from both the computer vision and natural language generation communities, \revise{and steady progress has been made due to the development of vision and language techniques~\citep{updown,hu2022scaling,mPLUG_2_2023,li2022blip,kuo2023haav}.}
% The task of image captioning has drawn much attention from both the computer vision and natural language generation communities, and steady progress has been made due to the development of vision and language techniques~\citep{NIC,spatt,updown,mrnn}. 
\revise{Image captioning has aided many applications, ranging from summarizing photo albums to tagging images online. A particularly noteworthy application is a mobile app\footnote{https://www.apple.com/accessibility/vision/} that vocalizes the content captured by a smartphone camera, serving as an invaluable resource for visually impaired people by describing the world around them.}

% Although automatic image caption generators can accurately describe the image, they generate generic captions for images with similar semantic meaning, lacking the intrinsic human ability to describe the unique details of a specific image to distinguish it from other images. For instance, as shown in Figure~\ref{fig:teaser_figure}, simply mentioning the traffic light without explaining the specific meaning~(\eg, the color of the traffic light) cannot help visually-impaired people to make a decision whether or not to cross the street. 
\revise{Automatic image caption generators, while accurate, often produce generic captions for semantically similar images, missing unique details that differentiate one image from another. For example, as depicted in Figure~\ref{fig:teaser_figure}, a mention of a traffic light without specifying its color provides insufficient information for visually impaired individuals to decide about crossing the street. 
A model that describes the distinctive contents of each image is more likely to highlight the truly useful information. We define the {\em distinctiveness} of a caption by its capacity to {\em identify and articulate the unique objects or context of the target image, thereby differentiating it from semantically similar images.} This paper aims to enhance image captioning models with the capability to produce distinctive captions. }
% In this paper, we consider the goal of endowing image captioning models to generate distinctive captions. Here we refer to the {\em distinctiveness} of a caption as its ability to {\em describe the unique objects or context of the target image so as to distinguish it from other semantically similar images.}}

% Most of the existing image captioning models aim to generate a caption that best describes the semantics of a target image. Distinctiveness, on the other hand, requires the caption to best match the target image among similar images, \ie, describing the distinctive parts of the target image. As pointed out in~\citep{17_wang2020compare,19_wang2019describing,18_luo2020analysis}, traditional captioning models that optimize the cross-entropy loss or reinforcement reward (usually CIDEr score) may lead to over-generic image captions. Some efforts have been made to generate {\em diverse} captions to enrich the concepts by employing conditional GAN~\citep{cgan1,3_cgan}, VAE~\citep{jain2017creativity,wang2017diverse} or reinforcement learning~\citep{wang2019towards,wang2020diversity}. 
\revise{Existing image captioning models predominantly focus on generating captions that accurately reflect the semantics of a target image. Distinctiveness, on the other hand, requires the caption to best match the target image among similar images, \ie, describing the distinctive parts of the target image. Research has highlighted that traditional captioning approaches, which often rely on optimizing cross-entropy loss or reinforcement rewards (typically the CIDEr score), tend to produce overly generic captions ~\citep{17_wang2020compare,19_wang2019describing,18_luo2020analysis}. Some efforts have been made to generate {\em diverse} captions to enrich the concepts by employing conditional GAN~\citep{cgan1,3_cgan}, VAE~\citep{jain2017creativity,wang2017diverse} or reinforcement learning~\citep{wang2019towards,wang2020diversity}. }
% However, improving diversity cannot guarantee distinctiveness, since rephrasing the expressions or enriching the vocabulary do not necessarily introduce novel and distinctive information. 
Several methods are proposed to improve the distinctiveness by contrastive learning~\citep{2_disccap,1_contrastive,12_li2020context}, where they either aggregate the contrastive image features with the target image feature, or apply the contrastive loss to suppress the estimated conditional probabilities of mismatched image-caption pairs~\citep{1_contrastive,2_disccap}. 
However, the distractors are either a group of images with scene graphs that partially overlaps with the target image~\citep{12_li2020context}, or randomly selected unmatched image-caption pairs~\citep{1_contrastive,2_disccap}, which are easy to distinguish.
\citet{4_self-retrieval} and \citet{9_PSST} introduce self-retrieval reward with contrastive loss, which requires the generated captions to retrieve the target image in a visual-language space. However, weighing too much on image retrieval could lead a model to repeat the distinctive words~\citep{17_wang2020compare}, which hurts caption quality.

\revise{In this work, to generate distinctive captions, we consider the hard negatives, \ie, \emph{similar images} that generally share similar semantics with the target image, and push the generated captions to clearly show the difference between the target image and these hard negative images. 
For instance, as shown in Fig.~\ref{fig:teaser_figure}, the generated captions should specify the different aspects of the target image (\eg, different light colors and context) compared with other images that share similar semantics.} 
To this end, we propose a differential distinctive memory attention module that puts high attention on distinctive objects detected in the target image \abcadd{but not in the similar images, and low attention on objects that are common among the target and similar images}. Specifically, object features in the target image with low similarity to object features in similar images are considered more distinctive, and thus receive a higher attention value. 
\abcadd{Our proposed attention mechanism is a plug-and-play module that works to extend existing transformer-based captioning models.}
%\abc{also discuss the new loss functions that we use for distinctive training.} 
\wenjiaadd{Moreover, we further propose two loss functions to facilitate the training, which encourages the model to focus its captions on the distinct image regions: 1)
%and speak them out. 
the memory classification loss predicts the distinctive words from the image features; 2) the weighted distinctive loss encourages the captioning model to predict distinctive words describing the unique image regions and gives higher weights to the distinctive words that are highly related to the image.}

In summary, the contributions of this paper are three-fold: 

\revise{1) We propose a Group-based Differential Distinctive Captioning Method (DifDisCap), which constructs \revise{memory features} from object regions, weighted by their distinctiveness within an image group, to generate captions that uniquely describe each image in the group. Specifically, our model employs a memory difference encoding technique designed to accentuate the feature differences between the target image and its corresponding group of similar images.}

% 2) To enforce the weighted memory to contain distinctive object information, we further propose two distinctive losses, where the supervision is the distinctive words occurring in the ground-truth (GT) captions. 

\revise{2) To ensure the weighted memory contains distinctive object information, we introduce two novel distinctive loss functions. These functions are supervised by the occurrence of distinctive words found in the ground-truth captions, thereby reinforcing the emphasis on unique object details within the images.}

\revise{3) We have carried out comprehensive experiments and user studies, which demonstrate that our proposed model is able to generate distinctive captions. Furthermore, our model emphasizes the unique regions of each image, enhancing the interpretability.}

%\jimmy{describe the difference between conference paper and journal paper}
%\jimmy{need to refine} 
The preliminary conference version of our work has been published in \citet{wang2021group}. This journal article extends our preliminary work in four aspects.  First, we propose a memory difference encoding to emphasize the feature difference between the target image and the similar image group. Second, we propose Indicated Training (IndTrain) to apply our memory attention only to those distinctive GT captions, which makes distinctive training more effective. 
%suppresses the confusion within training. 
\wenjiaadd{Third, in order to emphasize those distinctive words highly related to the target image and discards the unrelated words, we measure the text-image \emph{relatedness} with a pretrained multi-modal network~(\ie, CLIP~\citep{radford2021learning}) and weight the distinctive word loss~\citep{wang2021group} according to this \emph{relatedness}.} 
Moreover, using the newly introduced DifDisCap method, we present new state-of-the-art results on several baseline models.

The remainder of the paper is organized as follows. In Section~\ref{sec:Related}, we present related works, including image captioning models and metrics. In Section~\ref{sec:method}, we introduce our Group-based Differential Distinctive Captioning method. The experimental setting and quantitative results are presented in Section \ref{sec:Exp}, and the user study and qualitative results are presented in Sections \ref{sec:user} and \ref{sec:Quali}. Finally, we conclude the paper in Section~\ref{sec:conclude}.

%----------Related work--------------
\section{Related work}
\label{sec:Related}

\subsection{Image captioning} 
Image captioning bridges two domains---images and texts. Classical approaches usually extract image representations using a convolutional neural network~(CNN), then feed them into a recurrent neural network~(RNN) and output sequences of words~\citep{NIC,mrnn,karpathy2015deep}. Recent advances mainly focus on improving the image encoder and the language decoder. For instance, \citet{updown} propose bottom-up features, which are extracted by a pre-trained Fast R-CNN~\citep{renNIPS15fasterrcnn} and a top-down attention LSTM, where an object is attended in each step when predicting captions. Apart from using RNNs as the language decoder, some works~\citep{aneja2018convolutional,wang2018cnn+,wang2018gated} utilize CNNs since LSTMs cannot be trained in a parallel manner. More recently, some approaches~\citep{li2019entangled,cornia2020meshed} adopt transformer-based networks with multi-head attention to generate captions, which mitigates the long-term dependency problem in LSTMs and significantly improves the performance of image captioning. Recent advances usually optimize the network with a two-stage training procedure, where they pre-train the model with word-level cross-entropy loss~(XE) and then fine-tune with reinforcement learning~(RL) \abcadd{using the CIDEr score~\citep{CIDEr} as the reward}. Also, some work \citep{wang2020neighbours} introduces similar images to improve the accuracy of the generated captions. However, as pointed out in~\citet{17_wang2020compare,1_contrastive,3_cgan}, training with XE and RL may encourage the model to predict an “average” caption that is close to all ground-truth (GT) captions, thus resulting in over-generic captions that lack distinctiveness. 
%\abc{add a few sentences contrasting with our approach.} 
\wenjiaadd{In contrast, our work gives higher attention to the  image regions that are different from other similar images, leading to more distinctive  captions. We further propose weighted distinctive loss to encourage the model to predict distinctive words.}

The above models typically focus on improving the accuracy of generated captions. 
%Recently, there are novel image captioning models enriching the research topics.
\abcadd{Recently, various works aim to expand on traditional image captioning by better utilizing cross-domain and linguistic knowledge, linking words to objects in the image, and addressing dataset bias.}
\citet{zhao2020cross} and \citet{yuan2022discriminative} propose cross-domain image captioning models that are trained on a source domain and generalized to other domains, to alleviate the demands for massive data in target domains. Moreover, \citet{chen2022visualgpt} propose to adapt the linguistic knowledge from large pretrained language models such as GPT~\citep{radford2018improving} to image captioning models. LEMON~\citep{hu2022scaling}, mPLUG~\citep{mPLUG_2_2023} and BLIP-2~\citep{li2023blip2} leverage the visual and semantic information from the vision-language pretrained model to boost the performance of image captioning. Apart from cross-domain adaptation, other image captioning models aim to ground objects in images~\citep{zhou2020more,huang2020image} and 3D scenes~\citep{cai20223djcg}.
%are drawing much attention. 
\citet{jiang2022visual} propose to implicitly link the words in captions and the informative regions on images with a cluster-based grounding model. Furthermore, \citet{kuo2022beyond} combine attribute detection with image captioning to achieve accurate attention localization. Besides, understanding and quantifying the social biases in image captioning, \eg, gender bias~\citep{hirota2022quantifying}, racial bias~\citep{zhao2021understanding} and emotional bias~\citep{mohamed2022okay}, can inspire new directions for mitigating the biases found in image captioning datasets and evoke models with less bias.

More relevant to our work are the recent works on group-based image captioning~\citep{12_li2020context,13_vedantam2017context,11_chen2018groupcap}, where a group of images is utilized as context when generating captions. \citet{13_vedantam2017context} generate sentences that describe an image in the context of other images from closely related categories. \citet{11_chen2018groupcap} summarize the unique information of the target images contrasting to other reference images, and~\citet{12_li2020context} emphasize both the relevance and diversity. Our work is different in the sense that we simultaneously generate captions for each image in a similar group, and highlight the difference among them by focusing on the distinctive image regions \abcadd{and object-level features}. Both \citet{11_chen2018groupcap} and \citet{13_vedantam2017context} extract one image feature from the FC layer for each image, where all the semantics and objects are mixed up. While our model focuses on the object-level features and explicitly finds the unique objects that share less similarity with the context images, leading to fine-grained and concrete distinctiveness. 

\revise{To construct groups of images that share similar semantics, our methodology initially involves randomly selecting an image as the target. Subsequently, we retrieve its nearest images using a visual-semantic retrieval model. Visual-semantic retrieval models, predominantly based on a one-to-one mapping of instances into a shared embedding space, are well-suited to retrieve images with similar characteristics. One widely-adopted method involves maximizing the correlation between related instances within a shared embedding space, \eg, using canonical correlation analysis to maximize the correlation between images and text~\citep{rasiwasia2010new,yan2015deep}. Another popular approach is based on triplet ranking, which aims to ensure that the distance between positive image-text pairs is smaller than that between negative pairs. Drawing inspiration from hard negative mining, VSE++\citep{faghri2017vse++} leverages maximum violating negative pairs to enhance performance. More recently, CLIP~\citep{radford2021learning} introduced a visual-language model employing a contrastive learning objective across various image-text pairs, while ALIGN~\citep{jia2021scaling} expands this methodology by incorporating noisy text descriptions on a larger scale. Our work uses both VSE++ and CLIP for the construction of similar image groups and showcases the performance in the following sections.}

% Drawing inspiration from hard negative mining, VSE++\citep{faghri2017vse++} leverages maximum violating negative pairs to enhance performance. More recently, CLIP\citep{radford2021learning} introduced a visual-language model employing a contrastive learning objective across various image-text pairs, while ALIGN~\citep{jia2021scaling} expands this methodology by incorporating noisy text descriptions on a larger scale. Our work integrates both VSE++ and CLIP for the construction of similar image groups.

\revise{Zero-shot image captioning tasks aim to develop robust image captioning models that advance the state-of-the-art both in terms of accuracy and fairness. To rigorously assess the capabilities of image captioning models in a zero-shot context, a comprehensive evaluation dataset~\citep{kim2023nice} has been introduced. This dataset is designed to explore the full potential of image captioning models under zero-shot conditions and evaluate various zero-shot captioning approaches. Specifically, a retrieval-augmented zero-shot captioning model~\citep{kim2023nice} utilizes external contextual knowledge complementary to the knowledge in the original model, and consequently helps the captioner to achieve higher accuracy. Notably, the focus of our paper is not on addressing the task of zero-shot captioning. Instead, our work concentrates on enhancing the distinctiveness of the generated captions.}

Zero-shot image captioning endeavors seek to enhance image captioning models by advancing both accuracy and fairness, specifically by addressing societal biases. To rigorously assess the capabilities of image captioning models in a zero-shot context, a comprehensive evaluation dataset~\citep{kim2023nice} has been introduced. This dataset is designed to explore the full potential of image captioning models under zero-shot conditions and evaluate various zero-shot captioning approaches. Furthermore, the integration of a retrieval-augmented model leverages external contextual knowledge, augmenting the intrinsic knowledge of the original model. This strategic addition facilitates the generation of captions with improved accuracy, thereby pushing the boundaries of state-of-the-art in image captioning.

\subsection{Distinctive and diverse image captioning}
Distinctive image captioning aims to overcome the problem of generic image captioning, by describing sufficient details of the target image to distinguish it from other images. 
\citet{1_contrastive} promote the distinctiveness of an image caption by contrastive learning.
The model is trained to give a high probability to the GT image-caption pair and a low probability to a randomly sampled negative pair. 
\citet{2_disccap} and \citet{4_self-retrieval} take the same idea that the generated caption should be similar to the target image rather than other distractor images in a batch, and applies caption-image retrieval to optimize the contrastive loss. However, the distractor images are randomly sampled in a batch, which can be easily distinguished from the target images. In contrast, in our work, we consider {\em hard negative images} that share similar semantics with the target image, and push the captions to contain more details and clearly show the difference between these images. 

More recently, \citet{17_wang2020compare} propose to give higher weight to the distinctive GT captions during model training. \citet{11_chen2018groupcap}  model the diversity and relevance among positive and negative image pairs in a language model, with the help of a visual parsing tree \citep{2017StructCap}.
In contrast to these works, our work compares a group of images with a similar context, and highlights the unique object regions in each image to distinguish them from each other. 
That is, our model infers which object-level features in each image are unique among all images in the group. Our model is applicable to most of the transformer-based captioning models.

\abcadd{Additionally, several works aim to improve the diversity of generated captions, where the model can generate a set of different captions for the same image.}
One group of works is based on conditional GANs~\citep{3_cgan,cgan1} and auto-encoders~\citep{aneja2019sequential,mahajan2020diverse}. However, promoting the variability of generated captions may not improve the distinctiveness~\citep{wang2022distinctive}. For instance, using synonyms and changing the word orders in generated captions encourage diversity in syntax and word usage, but do not introduce distinctiveness information.

\revise{Our work is relevant to the object detection models that associate object regions in images with semantic labels. Deep learning detectors can be classified into two distinct categories. The first is the two-stage detectors, which encompass both region proposal and bounding box regression modules~\citep{he2015spatial}, as exemplified by Fast R-CNN, a widely recognized network in this category~\citep{renNIPS15fasterrcnn}. The second category comprises one-stage detectors that divide the image into regions and predict bounding boxes and probabilities for each region simultaneously, with YOLO being a prominent network in this category~\citep{redmon2016you}. Our work leverages spatial image features derived from Fast R-CNN to construct \revise{memory features} for object regions. These vectors are weighted based on their distinctiveness within the image group, enabling the generation of distinctive captions. }

\subsection{Metrics for image captioning}

In recent years, many metrics have been proposed to assess the performance of a captioning model, most of which evaluate the fluency and accuracy, e.g., BLEU~\citep{bleu}, CIDEr~\citep{CIDEr}, SPICE~\citep{spice}, and METEOR~\citep{Meteor}. 
However, these traditional metrics normally evaluate the word-level and phrase-level similarity between generated captions and the GT captions, instead of considering the semantic similarity~\citep{stefanini2022show}. Furthermore, the captioning models \abcadd{trained with reinforcement learning to} optimize these metrics~\citep{cornia2020meshed,2_disccap} tend to generate over-generic captions instead of pointing out the distinctive details in each image~\citep{wang2022distinctive}. Some related works propose diversity metrics to evaluate the corpus-level diversity. For instance,  \citet{van2018measuring} propose a metric to quantify the number of unique words in the captions, and further calculates the \abcadd{number of} unique bigram or unigrams appearing in the generated captions. \citet{wang2017diverse} measure the percentage of novel sentences that do not appear in the training set. ~\citet{wang2020diversity} employ latent semantic analysis to quantify the semantic diversity of generated captions. These metrics measure the variability of generated words and phrases, but cannot tell if the generated captions can distinguish the target image from other similar ones, \ie, the distinctiveness. 

The first metric for distinctiveness was the retrieval method, which employs a pretrained semantic-visual embeddings model VSE++~\citep{faghri2017vse++} to retrieve the target image with the generated captions and reports the Recall at k. % (R@k). 
Ideally, a distinctive caption should %be able to 
retrieve the correct image as the first item in the retrieval list.
Furthermore, \citet{wang2022distinctive} consider that distinct captions are less similar to other captions, where the similarity is measured by the CIDEr  between generated captions and the GT captions of other similar images. While these metrics only consider the sentence level distinctiveness, we argue that captions describing the unique details of target images usually contain distinct words. In this paper, we propose two novel metrics that consider both word-level and sentence-level distinctness.

% The metrics for image captioning evaluation have raised much attention recently, including the evaluation of zero-shot captioning and non-reference captioning. BERTScore~\citep{zhang2019bertscore} introduces an automatic evaluation metric for text generation by computing a similarity score using contextual embeddings for each token in the candidate sentence with each token in the reference sentence. ViLBERTScore~\citep{lee2020vilbertscore} further evaluates image caption using both image and text information by generating image-conditioned embeddings for each token. Considering the drawback of BERTScore that requires reference captions, UMIC~\citep{lee2021umic} further proposes a new metric that does not require reference captions to evaluate image captions via contrastive learning. To evaluate the zero-shot captioning models, V-METEOR~\citep{demirel2022caption} measures the visual and non-visual content of generated sentences separately.

\revise{Recent advancements in image captioning evaluation metrics have garnered significant attention, particularly in the domains of zero-shot and non-reference captioning metrics. BERTScore~\citep{zhang2019bertscore} introduces an automated metric for text generation that calculates similarity scores by utilizing contextual embeddings to compare tokens in candidate sentences against those in reference sentences. Extending this concept, ViLBERTScore~\citep{lee2020vilbertscore} enhances image caption evaluation by incorporating both textual and visual information, generating image-conditioned embeddings for each token. Addressing the limitation of BERTScore that requires reference captions, UMIC~\citep{lee2021umic}, PAC-S~\citep{sarto2023positive}, and CLIP-S~\citep{hessel2021clipscore} introduce innovative metrics that evaluate image captions without reference captions. Furthermore, to specifically assess zero-shot captioning models, V-METEOR~\citep{demirel2022caption} has been proposed, which evaluates the visual and textual content of generated sentences independently, providing a more nuanced analysis of caption quality. While the aforementioned metrics demonstrate commendable performance in evaluating captioning models in the absence of reference captions, our research diverges in focus. Specifically, we concentrate on assessing the distinctiveness of the generated captions.}

\subsection{Attention mechanisms}
Attention mechanisms apply visual attention to different image regions when predicting words at each time step, and have been widely utilized in image captioning~\citep{spatt,you2016image,chen2017sca,15_guo2020normalized,pan2020x}. For instance, \citet{you2016image} adopt semantic attention to focus on the semantic attributes in the image. \citet{updown} exploit object-level attention with bottom-up attention, then associates the output sequences with salient image regions via a top-down mechanism. More recently, self-attention networks introduced by \revise{Transformers~\citep{vaswani2017attention} are widely adapted in both language and vision tasks~\citep{dosovitskiy2020image,ye2019cross,ramachandran2019stand,yang2020bert,su2019vl,2_disccap}.} \citet{15_guo2020normalized} normalize the self-attention module in the transformer to solve the internal covariate shift. \citet{huang2019attention} weight the attention information by a context-guided gate.
These works focus on learning self-attention between every word token or image region in one image. 
\citet{12_li2020context} migrate the idea of self-attention to visual features from different images, and averages the group {\em image-level} vectors with self-attention to detect prominent features. In contrast, in our work, we take a further step by proposing learnable memory attention 
that highlights prominent {\em object-level} R-CNN features with distinct semantics among similar images.

%among the R-CNN {\em object-level} features~\cite{updown} extracted %from similar images, to highlight the prominent features that with distinct semantics 
%convey distinguishing semantics 
%in the similar image group.

\begin{figure*}[t]
	\begin{center}
		\includegraphics[width=\linewidth]{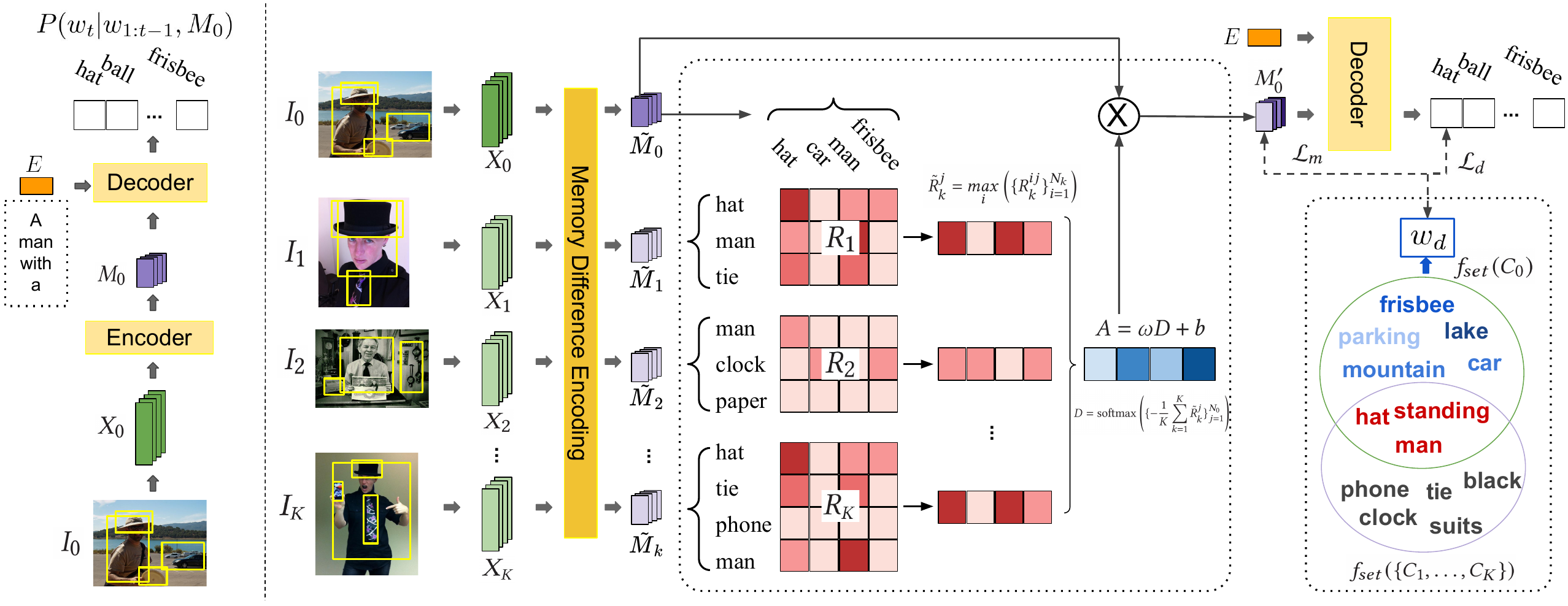}
	\end{center}
	\caption{Left: the standard transformer-based captioning model, where the target image features $X_0$ are the region-based visual features extracted via RoI pooling from Fast R-CNN. Right: our Group-based Differential Distinctive Captioning method (DifDisCap), which consists of a group-based differential memory attention (GDMA) module that weights the \revise{memory features} according to their similarity with other similar images. The words in blue are distinctive words $w_d$, and the words with higher relatedness are marked in the darker color. Our model takes a group of images as input, and outputs one caption for each image. Only one target memory $M_0'$, one decoder, and one output caption are shown here to reduce clutter. 
	}
	\label{fig:model}
\end{figure*}

%----------Methodology--------------
\section{Methodology}
\label{sec:method}

We present the framework of our proposed Group-based Differential Distinctive Captioning method (DifDisCap) in Figure~\ref{fig:model}. Our model aims to generate distinctive captions for each image within a group of semantically similar images. Given an image group with $K+1$ images, denoted as $\{I_0, I_1, \dots, I_K \}$,  DifDisCap generates distinctive captions for each image. 
Different from the conventional image captioning task, the generated captions should describe both the salient content in the target image, and also highlight the uniqueness of the target image (\ie, $I_0$) compared to other $K$ images (\ie, $I_1$ to $I_K$) in the same group. Specifically, during training, each image in the group is treated equally, and we use each image as a target iteratively. In Figure~\ref{fig:model}, we show an example where $I_0$ is the target image.

To achieve the goal of distinctive image captioning, we first construct similar image groups, comprising semantically similar images, then we employ the proposed Group-based Differential Memory Attention~(GDMA) module to extract the distinctive object features. Finally, we design two distinctive losses to further encourage generating distinctive words.

\subsection{Similar image group}

Similar image groups were first introduced in \citet{17_wang2020compare} to evaluate the distinctiveness of the image captions. For training, our model handles several similar image groups as one batch, simultaneously using each image in the group as a target image.
Here, we dynamically construct similar image groups during training as follows:

1) To construct a similar image group, we first randomly select one image as the target image $I_0$, and then retrieve its $K$ nearest images through the visual-semantic retrieval model VSE++~\citep{faghri2017vse++}, as in \citet{17_wang2020compare}. In detail, given the target image $I_0$, we use VSE++ to retrieve those captions that well describe $I_0$ among all human-annotated captions, and then the corresponding images of those captions are similar images.

2) Due to the non-uniform distribution of training images, the images sharing similar semantic meanings will form clusters in the VSE++ space. The images in the cluster center may be close to many other images, and to prevent them from dominating the training, the $K+1$ images that are used to create one similar image group are removed from the image pool in that, so they will not be selected when constructing other groups in the epoch. In this way, each image will belong to only one group, with no duplicate images appearing in one epoch.\footnote{When almost all images are selected, the remaining images are not similar enough to construct groups. We regard them as target images one by one, and find similar images from the whole image pool. }

Each data split (training, validation, test) is divided into similar image groups independently.
For each training epoch, we generate new similar image groups to encourage training set diversity.

\subsection{Group-based Differential Distinctive Captioning Method} 

Here we introduce the group-based differential distinctive captioning method (DifDisCap), and how we incorporate the Group-based Differential Memory Attention~(GDMA) module that encourages the model to generate distinctive captions. Notably, the GDMA can serve as a plug-and-play module for distinctive captioning, which can be applied to most existing transformer-based image captioning models. 

\subsubsection{Transformer-based Image Captioning}

Our captioning model is built on a transformer-based architecture~\citep{cornia2020meshed}, as illustrated in Figure~\ref{fig:model}~(left). The model can be divided into two parts: an image {\em Encoder} that processes input image features, and a caption {\em Decoder} that predicts the output caption word by word. In transformer-based architectures,  the {\em Encoder} and {\em Decoder} are both composed of several multi-head attention and MLP layers. 

In our work, we take the bottom-up features~\citep{updown} extracted by Fast R-CNN~\citep{renNIPS15fasterrcnn} as the input. Given an image $I$, let $X=\{x^i\}_{i=1}^N$ \revise{denote} the object features, where $N$ is the number of region proposals and $x^i \in \real^{d}$ is the feature vector for the $i$-th proposal. The output of the $l$-th encoder layer is calculated as follows:
\begin{align}
	O^{att}_{l} &= \textbf{LN}\left(X_{l-1} + \textbf{MH}\left(\mathbf{W_q}X_{l-1}, \mathbf{W_k}X_{l-1}, \mathbf{W_v}X_{l-1}\right)\right), \\
	X_{l} &= \textbf{LN}\left(O^{att}_l + \textbf{MLP}\left(O^{att}_l\right)\right),
\end{align}
where $\textbf{LN}(\cdot)$ denotes layer normalization, $\textbf{MLP}(\cdot)$ denotes a multi-layer perceptron, and $\textbf{MH}(\cdot)$ represents the multi-head attention layer. $\mathbf{W_q}, \mathbf{W_k}, \mathbf{W_v}$ are learnable parameters.

The {\em Encoder} turns features $X$ into \revise{memory features} $M =\{m^i\}_{i=1}^N$, where $m^i \in \real^{d_m}$ encodes the information from the $i$-th object proposal $x^i$, and is affected by other objects in the multi-head attention layers, which contains both single object features and the relationships among objects. According to the memory  $M$ and the embedding $E$ of the previous word sequence $\{w_1, \dots, w_{t-1} \}$, the {\em Decoder} generates the $v$-dimensional word probability vector $P_t = P(w_t|w_{1:t-1},M)$ at each time step $t$, where $v$ is the size of vocabulary.

\begin{figure}[t]
	\begin{center}
		\includegraphics[width=.9\linewidth]{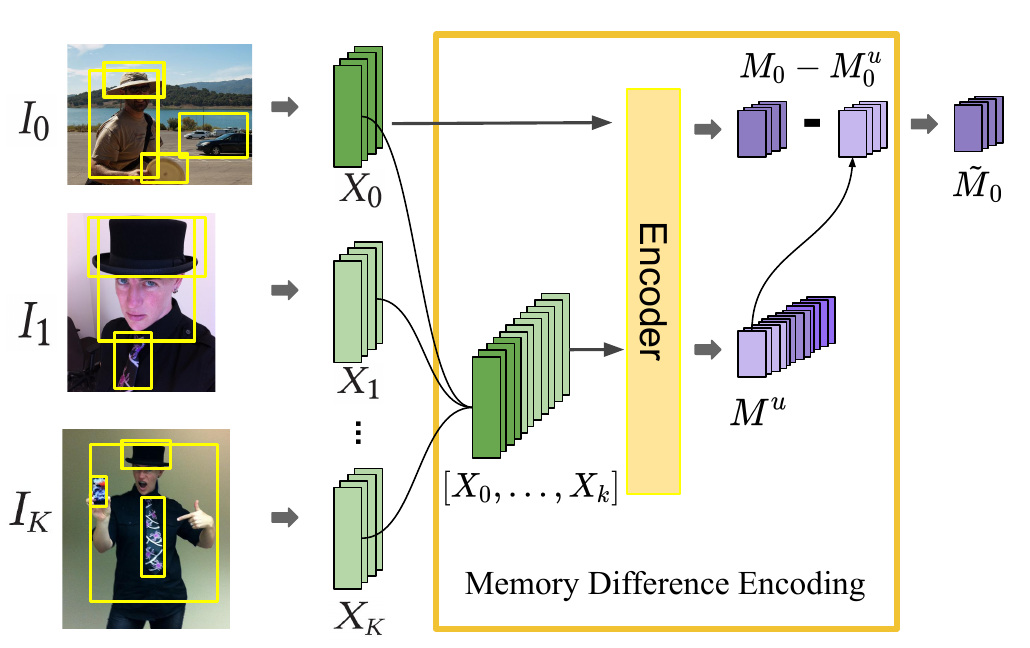}
	\end{center}
	\caption{The architecture of the Memory Difference Encoding module. The union memory vector $M_{u}$ encodes the fused information from images $I_0$ to $I_K$, and  $\tilde{M}_0$ encodes the difference between the target image $I_0$ and other similar images.
	%\abc{in the figure, change $M_0^d$ to $\tilde{M}_0$. Also in Fig 2.} \wenjia{Finished.}
	}
	\label{fig:mem_dif}
\end{figure}

\subsubsection{Group-based Differential Memory Attention (GDMA)}

The goal of group-based differential memory attention is to highlight the distinctive features of the target image that do not appear in other similar images. For instance, in Figure~\ref{fig:model}~(right), the concept of \textit{man} and \textit{hat} that appear in the target image $I_0$ are also shared in other similar images, but \textit{frisbee} and \textit{cars} are unique for $I_0$ and can distinguish $I_0$ from other images. However, the standard captioning model in Figure~\ref{fig:model}~(left) cannot highlight those objects, since each memory vector $m_0^i$ for different image regions is treated equally when fed into the {\em Decoder}.  Here we propose memory difference encoding and distinctive attention to highlight the distinct regions by assigning higher weights to their corresponding \revise{memory features}.

%\subsubsection{Memory Difference Encoding}
\myparagraph{Memory Difference Encoding.} 

\revise{Existing methods usually predict the caption $\hat{c}$ only according to the image $I_0$, while our model also consider $\{I_0, \dots, I_k\}$. The memory features $M_0$ is obtained via a feature encoder $f_{en}$:}
\begin{align}
	M_0 = f_{en}(X_0). \label{equ:m0}
\end{align}
In this work, we need the \revise{memory features} $\tilde{M}_0$ for image $I_0$ to contain the difference between $I_0$ and other similar images (i.e., $I_1, \dots, I_K$).  We therefore propose a Memory Difference Encoding module as shown in Figure~\ref{fig:mem_dif}. Instead of simply encoding $X_0$ to $X_K$ separately, we first concatenate the $K+1$ images feature (i.e., $X_0$ to $X_K$) 
and apply the encoder to obtain a union memory $M^u$, 
\begin{align}
	M^u = \abcadd{[M^u_0, \dots M^u_k]} = f_{en}([X_0, \dots, X_k]) \,. \label{equ:mu}
\end{align}
The union memory vector $M^u_0$ encodes the fused information of image $X_0$ and other similar images feature ($X_1$ to $X_k$), where $M_0^u$ is the part of $M^u$ corresponding to $X_0$. Next, we extract the difference between $M_0$ and the union memory vector $M_0^u$ as $M_0^d$:
\begin{align}
	\tilde{M}_0 = M_0 - M_0^u \,, \label{equ:m0u} 
\end{align}
and employ the difference encoding $\tilde{M}_0$ to generate the caption for $I_0$.

\myparagraph{Computing distinctive attention.}
In this work, we aim to give higher attention to the distinctive image regions when generating captions. Hence, the model will describe the distinctive aspects of the input image instead of only describing the most salient regions. 
To this end, we propose the group-based differential memory attention~(GDMA) module (see Figure~\ref{fig:model} (right)), where the attention weight for each object region is obtained by calculating the distinctiveness of its memory vector $\tilde{m}_0^i$. Then we encourage the model to generate distinctive words associated with the unique object regions.  
Specifically, the GDMA produces distinctive attention $A = \{a_i\}_{i=1}^{N_0} \in \real^{N_0}$ for differential \revise{memory features} $\tilde{M}_0 = \{\tilde{m}_0^1, \dots, \tilde{m}_0^{N_0}\}$. When generating captions, instead of using $\tilde{M}_0$, a weighted target memory is fed into the decoder:
\begin{align}
	M_0'=\{a_i \cdot \tilde{m}_0^i\}_{i=1}^{N_0} \,.
	\label{equ:weight}
\end{align}

To compute the distinctive attention, we need to compare the objects in the target image with those in similar images.
As shown in Figure~\ref{fig:model}~(right),  the target image $I_0$ and its similar images $\{I_k\}_{k=1}^K$ are transferred into 
\revise{memory features} $\tilde{M}_0=\{\tilde{m}_0^j\}_{j=1}^{N_0}$ and 
$\tilde{M}_k=\{\tilde{m}_k^i\}_{i=1}^{N_k}$ ($k=1,\dots,K$), via the image encoder, where $N_k$ denotes the number of objects in the $k$-th image. 
The GDMA first measures the similarity $R_k \in \real^{N_k \times N_0}$  of each target memory vector $m_0^j$ and each memory vector $m_k^i$ in similar images
via cosine similarity:
\begin{align}
	R^{ij}_k = \cos(\tilde{m}^i_k, \tilde{m}^j_0) \,,
	\label{eqn:R}
\end{align}
where $\tilde{m}_0^j \in \real^{d_m}$ is the $j$-th vector in $\tilde{M}_0$ (\eg, \revise{memory features} for \textit{hat}, \textit{car}, \textit{man} and \textit{frisbee} in Figure~\ref{fig:model}), and $\tilde{m}_k^i$ is the $i$-th vector in $\tilde{M}_k$ (\eg, \revise{memory features} for \textit{hat}, \textit{man}, and \textit{tie} from $\tilde{M}_k$ in Figure~\ref{fig:model}). 

The similarity matrix reflects how common an object is---a common object that occurs in many images is not distinctive for the target image. For example, as shown in Figure~\ref{fig:model} (right), \textit{hat} is less distinctive since it occurs in multiple images, while \textit{car} is an unique object that only appear in target image. 
To summarize the similarity matrix, we compute an object-image similarity map $\tilde{R}_k \in \real^{N_0}$ as
\begin{align}
%	\tilde{R}_k^j = \mathop{\max}\limits_{i} \left ( \{{R_k^{ij}}\}_{i=1}^{N_k} \right)\,, \\
	\abcadd{\tilde{R}_k^j = \mathop{\max}\limits_{i\in\{1,\dots,N_k\}} R_k^{ij} \,,}
\label{equ:r_tilde}
\end{align} 
where $\tilde{R}_k^j$ is the similarity of the best matching object-region in image $I_k$ to region $j$ in the target image $I_0$.

We assume that objects with higher similar scores are less distinctive. 
\abcadd{Hence we define the raw distinctiveness score for the $j$-th region in $I_0$, corresponding to memory vector $\tilde{m}_0^j$, as the negative average of its object-image similarity maps with the similar images,
\begin{align}
d_j = -\frac{1}{K}\sum_{k=1}^{K}{\tilde{R}_k^j}, \,\, j\in\{1,\dots,N_0\}.
\end{align}}
Here, higher scores indicate higher distinctiveness, i.e., lower average similarity to other similar images.
\abcadd{The raw distinctiveness scores are normalized by applying the softmax function, yielding the final distinctiveness scores $D \in \real^{N_0}$ for the \revise{memory features} $\{\tilde{m}_0^j\}$,}
%Hence, the distinctiveness scores $D \in R^{N_0}$ for each memory vector $\tilde{m}_0^j$ (j-th object region in the target image) are computed by softmax of the negative object-image similarity maps averaged over the similar images:
\begin{align}
	%D = \mathrm{softmax} \big(\big[-\frac{1}{K}\sum_{k=1}^{K}{\tilde{R}_k^j}\big]_{j=1}^{N_0} \big)\,,
	D = [D_1,\cdots, D_{N_0}] = \mathrm{softmax} \big([d_1,\cdots,d_{N_0}]\big)\,,
\end{align}
%where higher scores indicates higher distinctiveness, i.e., lower average similarity to other similar images.
Note that the values of $D$ range in $[0,1]$ due to the softmax function.

\myparagraph{Indicated Training.} In the training set, each target image $I_0$ has several ground truth (GT) captions. Not all these GT captions are distinctive, and thus attending to unique distinctive features may confuse the training process. We therefore use a distinctive metric (\eg, CIDErBtw~\citep{17_wang2020compare}) to divide the ground truth captions into \emph{distinctive} ones and \emph{common} ones. 

For those GT captions indicated as \emph{distinctive}, the distinctive attention weights $A=[a_1,\cdots,a_{N_0}]$ for the target \revise{memory features} in (\ref{equ:weight}) are calculated as:
\begin{align}
	A&=\omega D + b \,,
\label{equ:attention}
\end{align}
where $\omega$ and $b$ are two learnable parameters. The bias term $b$ controls the minimum value of $A$, i.e., the base attention level for all regions, while $\omega$ controls the amount of attention increase due to the distinctiveness. We clip $\omega$ and $b$ to be non-negative, so that the attention values in A are non-negative.

For those GT captions indicated as \emph{common}, we let $A = \boldsymbol{1}$ so that the \revise{memory features} are all considered equally in Equation~\ref{equ:weight}. 

%could also be well treated. $A$ is then used to highlight the distinctive memory vectors as in Equation~\ref{equ:weight}.

% \jimmy{Here Indicated Training (IndTrain)}

\subsection{Loss functions}

Two typical loss functions for training image captioning are cross-entropy loss and reinforcement loss. 
\wenjiaadd{The reinforcement loss uses \abcaddd{the average CIDEr with} the GT captions of the image for supervision, which may encourage the generated captions to mimic ``average'' GT caption, resulting in over-genericness, \ie, lack of distinctiveness}.
To address this issue, we take a step further to define distinctive words and explicitly encourage the model to learn more from these words. 
In this section, we first review the two typical loss functions used in captioning models, and then present our proposed weighted distinctive loss (WeiDisLoss) \abcadd{and memory classification loss (MemClsLoss)} for training our GDMA module.

\subsubsection{Cross-Entropy Loss}

Given the $i$-th GT caption of image $I_0$, $C^i_0=\{w_t\}_{t=1}^T$, the cross-entropy loss is 
\begin{align}
	\mathcal{L}_{xe}=  - \sum\limits_{t = 1}^T \log  P(w_t|w_{1:t-1}, M_0')\,,
\end{align}
where $P(w_t|w_{1:t-1}, M_0')$ denotes the predicted probability of the word $w_t$ conditioning on the previous words $w_{1:t-1}$ and the weighted \revise{memory features} $M_0'$, as generated by the caption {\em Decoder}.

\subsubsection{Reinforcement learning loss}

Following~\citet{rennie2017self}, we apply reinforcement learning to further improve the accuracy of our trained network using the loss:
\begin{align}
	\mathcal{L}_{r} = -E_{\hat{c}\sim p(c|I)} \left[\frac{1}{d_c}\sum\limits_{i = 1}^{d_c} g(\hat{c},C_0^i)\right],
	\label{eqn:newreward}
\end{align}
where $g(\hat{c},C_0^i)$ is the CIDEr value between the predicted caption $\hat{c}$ and the $i$-th GT $C_0^i$, and $d_c$ denotes the number of GT captions.

\subsubsection{Weighted distinctive loss (WeiDisLoss)} 

We propose a weighted distinctive loss to encourage the caption model to focus on the distinctive words that appear in captions $C_0$ of the target image, but not in captions $\{C_1, \dots, C_K\}$ of similar images. We define the distinctive word set $\Omega$ for $I_0$ as
%\abc{changed $w_d$ to $\Omega$ to avoid confusion with $w_t$. Use $\omega_i$ for a word in $\Omega$}
\begin{align}
	\Omega = f_{set}(C_0) - f_{set}(\{C_1, \dots, C_K\}) \,,
\end{align}
where $f_{set}(\cdot)$ denotes the function that converts the sentence into a word set, and ``$-$'' here means set subtraction.

In the training phase, we explicitly encourage the model to predict the distinctive words in $\Omega$ by optimizing the distinctive loss ${\mathcal L}_{d}$, 
\begin{align}
	{\mathcal L}_{d} = - \sum\limits_{t = 1}^T \sum\limits_{i = 1}^{|\Omega|}  \lambda_{\omega_i} \log P(w_t = \omega_i | w_{1:t-1}, M_0')  \,, \label{equ:disword}
\end{align}
where $\omega_i$ denotes the $i$-th distinctive word in $\Omega$, and $P(w_t = \omega_i | w_{1:t-1}, M_0')$ denotes the probability of predicting word $\omega_i$ as the $t$-th word in sentence. $|\Omega|$ is the number of words in $\Omega$, and $T$ is the length of the sentence.

In practice, due to the personalized language preference of each annotator, not every word in $\Omega$ is highly related to the image $I_0$, and those unrelated words would distract the captioning model.  We therefore apply a weight ${\lambda}_{\omega_k}$  to each term in the WeiDisLoss in (\ref{equ:disword}).
%measure the \emph{relatedness}
%\abc{``relatedness'' sounds clearer than "relativeness"}
%of the $k$-th distinct word $\omega_k \in \Omega$ with the target image $I_0$, which is indicated as ${\lambda}_{\omega_k}$, to weight the DisWrdLoss \abc{should it be WeiDisLoss?} in (\ref{equ:disword}). %
The weight ${\lambda}_{\omega_k}$  measures the \emph{relatedness}
%\abc{``relatedness'' sounds clearer than "relativeness"}
of the $k$-th distinct word $\omega_k$ with the target image $I_0$,
In detail, $\omega_k$ is placed into a sentence $c_{\omega_k}$ with the template ``this picture includes $\omega_k$'',  and the relatedness ${\lambda}_{\omega_k}$ is calculated as
\begin{align}
	\hat{\lambda}_{\omega_k} =\theta(c_{\omega_k})\cdot \phi(I_0)  \,, \\
	\lambda_{\omega_k} = \frac{\hat{\lambda}_{\omega_k}}{\max_{k} \hat{\lambda}_{\omega_k} } 
\end{align}
where $\theta(\cdot)$ and $\phi(\cdot)$ denote the sentence embedding and image embedding of a multimodal embedding model (\eg, CLIP~\citep{radford2021learning}).
\abcadd{Thus, distinctive words that are more related to the target image (according to the embedding model) will have higher weights in the loss.}

\revise{Since CLIP is weak at identifying small or particular objects, employing CLIP to assess the relevance of these distinct words to the target image might result in lower weighting for smaller objects.
This concern can be partly addressed by combining other loss functions that focus on small objects. For example, memory classification loss introduced in Section~\ref{sec:MemClsLoss} encourages the model to pay attention to all objects denoted by distinct words. Additionally, both the cross-entropy loss and the reinforcement learning loss encourage the model to generate captions following human supervision, regardless of the object sizes within these captions.}

% \jimmy{Here "WeightedDistinctWords"}

% \myparagraph{Distinctive Word loss (DisWrdLoss).} In the training phase, we explicitly encourage the model to predict the distinctive words in $w_{d}$ by optimize the distinctive loss ${\mathcal L}_{d}$, 
% \begin{align}
% 	{\mathcal L}_{d} = - \sum\limits_{t = 1}^T \sum\limits_{i = 1}^u  \lambda_{w_d^i} \log P(w_t = w_{d}^i | w_{1:t-1}, M_0')  \,,
% \end{align}
% where $w_{d}^i$ denotes the $i$-th distinctive word in $w_{d}$, and $P(w_t = w_{d}^i | w_{1:t-1}, M_0')$ denotes the probability of predicting word $w_{d}^i$ as the $t$-th word in sentence. $u$ is the number of words in $w_{d}$, and $T$ is the length of the sentence.

\subsubsection{Memory classification loss (MemClsLoss)} 
\label{sec:MemClsLoss}
%\abc{In this section, should "GMA" be "GDMA"?} \wenjia{Yes, revised.}
In order to generate distinctive captions, the {\em Decoder} requires the GDMA to produce memory contents containing distinctive concepts. However, the supervision of the GDMA through the {\em Decoder} could be too weak, which may allow the GDMA to also produce non-useful information, \eg, highlighting too much background or focusing on small objects that are not mentioned in the GT captions.
To improve the distinctive content produced by the GDMA, we introduce an {\em auxiliary classification task} that predicts the distinctive words from the weighted \revise{memory features} $M_0'$ of the GDMA, 
\begin{align}
	P_M = f_{MC}(M_0') \,,
\end{align}
where $P_M$ denotes the word probability vector and $f_{MC}$ is the classifier.
To associate the \revise{memory features} with distinctive words, we employ the multi-label classification loss ${\mathcal L}_{m}$ to train the classifier,
\begin{align}
	{\mathcal L}_{m} =  - \sum\limits_{k = 1}^{|\Omega|}  \lambda_{\omega_k} {\log (P_{M, \omega_k}) }\,,
\end{align}
where $P_{M, \omega_k}$ is the predicted probability of the $k$-th distinctive word.

\subsubsection{The final loss}

The final training loss ${\mathcal L}$ is formulated as 
\begin{align}
	{\mathcal L} = \alpha_{c}{\mathcal L}_{xe} + \alpha_{r}{\mathcal L}_{r} + \alpha_{d}{\mathcal L}_{d} + \alpha_{m}{\mathcal L}_{m} \,,
\end{align}
where $\{\alpha_{c}, \alpha_{r}, \alpha_{d}, \alpha_{m}\}$ are hyper-parameters for their respective losses.
\revise{The training procedure has two stages following~\citet{2_disccap}.}
In the first stage, we set $\alpha_{c} = 1$ and $\alpha_{r} =0$, so that the network is mainly trained by cross-entropy loss $\alpha_{c}$.  In the second stage, we set $\alpha_{c} = 0$ and $\alpha_{r} =1$, so that the parameters are mainly optimized by reinforcement learning loss ${\mathcal L}_{r}$.
We adaptively set $\{\alpha_d, \alpha_m\}$ so that
$\alpha_{d} {\mathcal L}_{d}$ and $\alpha_{m} {\mathcal L}_{m}$ are one quarter of ${\mathcal L}_{xe}$ (or ${\mathcal L}_{r}$).

During training, each mini-batch comprises several similar image groups, with the loss aggregated over each image as a target in its group. We show the details for processing one image group in Algorithm~\ref{algo:2}. 
%During training, each mini-batch comprises several similar image groups, with the loss aggregated over each image as a target in its group.

\renewcommand{\algorithmicrequire}{\textbf{Input:}}
\renewcommand{\algorithmicensure}{\textbf{Output:}}
\begin{algorithm} % begin
	\caption{The training procedure of DifDisCap in each step} % title
	\label{alg} % label
	\begin{algorithmic}[1] % number
		\Require A similar image group ${I_0,\dots, I_K}$ with captions ${C_0,…,C_{K}}$ % input
		\Ensure The final loss $ \mathcal{L}$ of this similar image group to optimize the {\em Encoder} and {\em Decoder} % output
		
		\State Encode the image group $\{ {I_0,\dots, I_K} \}$ into $\{M_0,\dots,M_K\}$, where the memory of  the $k$-th image is $M_k = \{m_k^i\}_{i=1}^{N_k}$
		\State Encode the whole image group into $[M^u_0,\dots,M^u_k] = f_{en}([X_0, \dots, X_k])$
		
		\For{$k \gets 0$ to $K$} 
		\State Calculate the memory difference $\tilde{M}_k = M_k - M_k^u$. 
		%, where $M_k^u$ is the part of $M^u$ corresponding to $X_k$.
		\State Calculate the distinctive attention $A$ as in (\ref{equ:attention}) for the distinctive ground-truth captions, where $A=\{ a_i \}_{ i=1}^{N_k} \in \real^{N_k}$.  Let $A = \boldsymbol{1}$ for the common ground-truth captions.
		%\abc{$f_{GDMA}$ is not defined anywhere. You can refer to the equation numbers instead.}\jimmy{Revised}
		\State Calculate weighted target memory: %Weight $M_k^d = \{m_k^1, \dots, m_k^{N_0}\}$  into $M_k’$: 
		$M_k’=\{ a_i \cdot \tilde{m}_k^i\}_{i=1}^{N_k}$
		\State Calculate distinctive word set $\Omega_k = f_{set}(C_k) - f_{set}(\{C_j\}_{j\neq k})$
		%\{C_i | i \in [1, K], i \in Z, i \neq k\})$
		\State Decode $M_k’$ as probability of generated words $\{P_t\}_{t=1}^T$
		\State Classify $M_k’$ as possible words $P_M \gets f_{MC}(M_k')$
		\State Calculate each loss for the $k$-th image, including: 
		\State \quad Cross-entropy loss $\mathcal{L}_{xe}$ with $P_t$ and $C_k$, 
		\State \quad Reinforcement learning loss $\mathcal{L}_c$ with $P_t$ and $C_k$,
		\State \quad Weighted distinctive word loss $ \mathcal{L}_d$ with $P_t$ and $\Omega_k$, 
		\State \quad Memory classification loss $\mathcal{L}_m$ with $P_M$ and $\Omega_k$.
		\State Get the loss for the $k$-th image $\mathcal{L}_k = \alpha_{c} \mathcal{L}_{xe} +  \alpha_{r} \mathcal{L}_r +  \alpha_{d} \mathcal{L}_d +  \alpha_{m} \mathcal{L}_m$
		\EndFor
		\State Accumulate the loss $\mathcal{L} = \sum\limits_{k=0}^K \mathcal{L}_k $
		
	\end{algorithmic}
	\label{algo:2}
\end{algorithm}

\section{Experiments}
\label{sec:Exp}

% main result table
\begin{table*}[tb]
	\begin{center}
        \resizebox{\linewidth}{!}{
			\begin{tabular}{c|ccc|ccc}
				\textbf{Method} &\textbf{DisWordRate(\%)$\uparrow$}& \textbf{CIDErRank$\downarrow$} & \textbf{CIDErBtw$\downarrow$} & \textbf{CIDEr$\uparrow$}   &  \textbf{BLEU3$\uparrow$} & \textbf{BLEU4$\uparrow$} \\
				\hline
				Transformer~\citep{2_disccap} & 16.8 & 2.47 & 74.8 & 111.7  & 45.1 & 34.0 \\
				\hspace{2mm}  + DifDisCap (ours) & \textbf{19.7} & 2.40 & \textbf{70.7} & 107.1 & 43.3 & 32.6 \\
				\hline
				M$^2$Transformer~\citep{cornia2020meshed}* & 16.4  & 2.52 & 76.8 &  111.8   & 45.2 & 34.7 \\
				\hspace{2mm}  + DifDisCap (ours)  & 18.8 & 2.43 & 72.4 & 109.8 & 44.3 & 33.5 \\
    
				\hline
				Transformer + SCST~\citep{2_disccap} & 14.7 & 2.38 & 83.2 & 127.6 & \textbf{51.3} & \textbf{38.9} \\
				\hspace{2mm}  + DifDisCap (ours) & 17.0 & 2.34 & 81.5 & 126.9 & 50.6 & 38.4 \\
    % \hspace{8mm}  \revise{+ RegionCLIP (ours)}& 17.3 & \textbf{2.20} & 81.7 & 128.1 & 51.1 & 38.7 \\
				\hline
				M$^2$Transformer + SCST~\citep{cornia2020meshed}* & 17.3 & 2.38 & 82.9  & \textbf{128.9}  & 50.6 & 38.7 \\
				\hspace{2mm}  + DifDisCap (ours)  & 18.7 & \textbf{2.30} & 80.1 & 127.2 & 49.9 & 38.2  \\
    % \hspace{8mm}  \revise{+ RegionCLIP (ours)} & 18.9 & 2.22 & 80.2 & 128.2 & 50.8 & 38.8 \\
				\hline
				FC~\citep{rennie2017self} & 6.5 & 3.03 & 89.7 & 102.7 & 43.2 & 31.2 \\
				Att2in~\citep{rennie2017self} & 10.8 & 2.65 & 88.0 & 116.7   & 48.0 & 35.5 \\
				UpDown~\citep{updown} & 12.9 & 2.55 & 86.7 & 121.5   & 49.2 & 36.8 \\
				AoANet~\citep{huang2019attention}* & 14.6 & 2.47 & 87.2 &128.6 & 50.4 & 38.2 \\
				\hline
				DiscCap~\citep{2_disccap} & 14.0 & 2.48 & 89.2 & 120.1 & 48.5 & 36.1 \\
				CL-Cap~\citep{1_contrastive} & 14.2 & 2.54 & 81.3 & 114.2 & 46.0 & 35.3 \\
			 CIDErBtwCap~\citep{17_wang2020compare} & 15.9 & 2.39 & 82.7 & 127.8  & 51.0 & 38.5 \\
		\end{tabular}}
		\caption{Comparison of caption distinctiveness and accuracy on MSCOCO test split: \textbf{DisWordRate}, \textbf{CIDErRank}, and \textbf{CIDErBtw} measure the distinctiveness, while \textbf{CIDEr} and \textbf{BLEU-3} and \textbf{BLEU-4} measure the accuracy. $\uparrow$ and $\downarrow$ show whether higher or lower scores are better according to each metric. We apply our model on four baseline models: Transformer~\citep{2_disccap} and M$^2$Transformer~\citep{cornia2020meshed} trained only with cross-entropy loss, Transformer + SCST~\citep{2_disccap} and M$^2$Transformer + SCST~\citep{cornia2020meshed} trained with reinforcement learning. * denotes we train the model from scratch with officially released code. 
		}
		\label{table:main_results_shrink}
	\end{center}
\end{table*}

In this section, we first introduce the implementation details and dataset preparation, then we quantitatively evaluate the effectiveness of our model through an ablation study and a comparison with other state-of-the-art models.

\subsection{Implementation details}

Following~\citet{updown}, we use the spatial image features extracted from Fast R-CNN~\citep{renNIPS15fasterrcnn} with dimension $d=2048$. Each image usually contains around $50$ object region proposals, i.e. $N_0 \approx 50$. Each object proposal has a corresponding memory vector with dimension $d_m = 512$. We set $K=5$ for constructing similar image groups. The values in \abcadd{learned} distinctive attention $A$ are mostly in the range of $(0.4, 0.9)$. To verify the effectiveness of our model, we conduct experiments on four baseline methods~(\ie, Transformer~\citep{2_disccap}, M$^2$Transformer~\citep{cornia2020meshed}, Transformer + SCST~\citep{2_disccap} and M$^2$Transformer + SCST~\citep{cornia2020meshed}). 
Our experimental settings (\eg, data preprocessing and vocabulary construction) follow these baseline models. We apply our GDMA module to the Transformer model and all three layers in M$^2$Transformer model. Note that our model is applicable to most of the transformer-based image captioning models, and we choose these four models as the baseline due to their superior performance on accuracy-based metrics.

\subsection{Dataset and metrics}

\subsubsection{Dataset}
We conduct the experiments using the most popular dataset---MSCOCO\footnote{https://cocodataset.org/\#download}, which contains 12,387 images, and each image has $5$ human annotations. Following \citet{updown}, we \wenjiaadd{utilize the Karpathy split~\citep{karpathy2015deep} to} divide the dataset into three sets---5,000 images for validation, 5,000 images for testing, and the rest for training.
%\abc{is this the Karpathy split?}\wenjia{Yes, revised.}.
When constructing similar image groups for the test set, we adopt the same group split as~\citet{17_wang2020compare} for a fair comparison.

\abcaddd{Note that we did not run our main experiments on the online MSCOCO test set since we need to evaluate distinctiveness using the below metrics, which are not available from the MSCOCO submission server. Please see the Appendix for the evaluation on the online MSCOCO test set using standard metrics, e.g., CIDEr.}
%\abc{do we perform tests on the online MSCOCO test set?} \jimmy{I have done online testing and reported the results on appendix. }

\subsubsection{Metrics}
We consider two groups of metrics for evaluation. 
The first group includes the metrics that evaluate the accuracy of the generated captions, such as CIDEr and BLEU.
The second group assesses the distinctiveness of captions.
For the latter, CIDErBtw~\citep{17_wang2020compare} calculates the CIDEr value between generated captions and GT captions of its similar image group. However, CIDErBtw only works when comparing two methods with similar CIDEr values, \eg, a random caption that has lower overlap with the GT captions will be considered distinctive since it achieves lower CIDErBtw. Hence, we propose two new distinctiveness metrics as follows. 

\myparagraph{CIDErRank.} A distinctive caption $ \hat{c}$ (generated from image $I_0$) should be similar to the target image's GT captions $C_0$, while different from the GT captions of other images in the same group $\{C_1, \dots, C_K\}$. Here we use CIDEr values $\{{s_k}\}_{k=0}^K$ to indicate the similarity of the caption $\hat c$ with GT captions of images in the same group as
\begin{align}
	s_k&=\frac{1}{d_c}\sum\limits_{i = 1}^{d_c} g(\hat{c},C_k^i)\,,
\end{align}
where $g(\hat{c},C_k^i)$ represents the CIDEr value of predicted caption $\hat{c}$ and $i$-th GT caption in $C_k$.
\abcadd{We sort the CIDEr values $\{{s_k}\}_{k=0}^K$ in descending order, and use the rank $r$ of $s_0$ to measure the distinctiveness of $\hat{c}$.}
%
%We use the rank of $s_0$ in $\{{s_k}\}_{k=0}^K$ to show the distinctiveness of the models as
%\begin{align}
%	r&=f_{rank} \left (s_0, \{{s_k}\}_{k=0}^K \right )\,,
%\end{align}
%where $f_{rank}(\cdot)$ means $s_0$ is the $r$-th largest %value in $\{{s_k}\}_{k=0}^K$.
The best rank is $r=1$, indicating the generated caption $\hat{c}$ is \abcadd{more similar to its GT captions 
the other captions}, 
%than different from other captions, 
while the worst rank is $K+1$.
Thus, the average $r$ reflects the performance of captioning models, with more distinctive captions having lower CIDErRank.

\myparagraph{DisWordRate.} We design this metric based on the assumption that using distinctive words should indicate that the generated captions are distinctive. The {\em distinctive word rate} (DisWordRate) of a generated caption $\hat{c}$ is calculated as:
\begin{align}
	DisWordRate= \max_i \frac{{|\Omega \cap \hat{c} \cap C_0^i|}}{{|\Omega \cap C_0^i|}}, \quad i=1,\dots, d_c \,,
\end{align}
\abcadd{where $\Omega$ is the set of distinctive words for image $I_0$,}
$d_c$ is the number of sentence in $C_0$, $|\Omega \cap \hat{c}  \cap C_0^i|$ represents the number of words in $\Omega$ that appear in both $\hat{c}$ and  $C_0^i$.
Thus, DisWordRate reflects the percentage of distinctive words in the generated captions.

\subsection{Main results}

\begin{table*}[tb]
    \begin{center}
	\begin{tabular}{l|ccc|ccc}
		\textbf{Method} &\textbf{DisWordRate(\%)$\uparrow$}& \textbf{CIDErRank$\downarrow$} & \textbf{CIDErBtw$\downarrow$} & \textbf{CIDEr$\uparrow$}   &  \textbf{BLEU3$\uparrow$} & \textbf{BLEU4$\uparrow$} \\
		\hline
		M$^2$Transformer & 16.4 & 2.52 & 76.8 & \textbf{111.8}   & \textbf{45.2} &  34.7  \\
		\hspace{2mm}+ ImageGroup & 16.9 & 2.51 & 75.4  &110.7   & \textbf{45.2} & \textbf{34.8} \\
		\hspace{4mm}+ WeiDisLoss& 17.3 & 2.48 & 74.3 & 110.0 & 45.0 & 34.4 \\
		\hspace{6mm} + GDMA& 18.6 & 2.45 & 72.7 & 110.1 & 45.1 & 34.6 \\
		\hspace{8mm} + IndTrain & \textbf{18.8} & \textbf{2.43} & \textbf{72.4} & 109.8 & 44.3 & 33.5 \\
		\hline
		M$^2$Transformer+SCST & 17.3 & 2.38 & 82.9 & \textbf{128.9} & \textbf{50.6} & \textbf{38.7} \\
		\hspace{2mm}+ ImageGroup & 17.4 & 2.36 & 82.3 & 127.8 & 50.4 & 38.5 \\
		\hspace{4mm}+ WeiDisLoss & 18.3 & 2.31 & 81.0 & 128.2 & 50.2 & 38.4 \\
		\hspace{6mm}+ GDMA& 18.4 & 2.31 & 80.4 & 127.6 & 49.9 & 38.3 \\
		\hspace{8mm}+  IndTrain & \textbf{18.7} & \textbf{2.30} & \textbf{80.1} & 127.2 & 49.9 & 38.2
	\end{tabular}
	\caption{Ablation study on four components. We train two baselines (\ie, M$^2$Transformer and  M$^2$Transformer+SCST), and gradually add four components of our full model: ImageGroup (image group based training), WeiDisLoss (weighted distinctive loss), GDMA(group-based differential memory attention), and IndTrain (Indicated Training). 
	}
	\label{table:ablation_shrink}
	\end{center}
\end{table*}

\begin{table*}[tb]
\resizebox{\textwidth}{13mm}{
\begin{tabular}{cc|ccc|ccc}
\textbf{Method} & \textbf{Region Feature} & \multicolumn{1}{l}{\textbf{DisWordRate} $\uparrow$} & \multicolumn{1}{l}{\textbf{CIDErRank}$\downarrow$} & \multicolumn{1}{l|}{\textbf{CIDErBtw}$\downarrow$} & \multicolumn{1}{l}{\textbf{CIDEr}$\uparrow$} & \multicolumn{1}{l}{\textbf{BLEU3}$\uparrow$} & \multicolumn{1}{l}{\textbf{BLEU4}$\uparrow$}  \\ \hline
\multirow{3}{*}{Transformer+SCST+DifDisCap} & Bottom-up & 17.0 & 2.34 & \textbf{81.5} & 126.9 & 50.6 & 38.4 \\
 & ViLBERT & 16.8 & 2.26 & 82.2 & 127.5 & 51.0 & 38.6 \\
 & RegionCLIP & \textbf{17.3} & \textbf{2.20} & 81.7 & \textbf{128.1} & \textbf{51.1} & \textbf{38.7} \\ \hline
\multirow{3}{*}{M$^2$Transformer+SCST+DifDisCap} & Bottom-up & 18.7 & 2.30 & \textbf{80.1} & 127.2 & 49.9 & 38.2 \\
 & ViLBERT & 18.3 & 2.24 & 82.1 & 127.8 & 50.3 & 38.4 \\
 & RegionCLIP & \textbf{18.9} & \textbf{2.22} & 80.2 & \textbf{128.2} & \textbf{50.8} & \textbf{38.8}
\end{tabular}
}
\caption{\revise{Comparison of different kinds of region feature, i.e., Bottom-up~\citep{updown}, ViLBERT~\citep{lu2019vilbert}, and RegionCLIP~\citep{zhong2022regionclip}. We experiment on two models (i.e., Transformer+SCST+DifDisCap and M$^2$Transformer+SCST+DifDisCap).}}
\label{tab:region_feature}
\end{table*}

In the following, we present a comparison with the state-of-the-art (distinctive) image captioning models. In addition, we present the ablation studies of our model.

% trade-off figure
\begin{figure}
	\begin{center}
		\includegraphics[width=.95\linewidth]{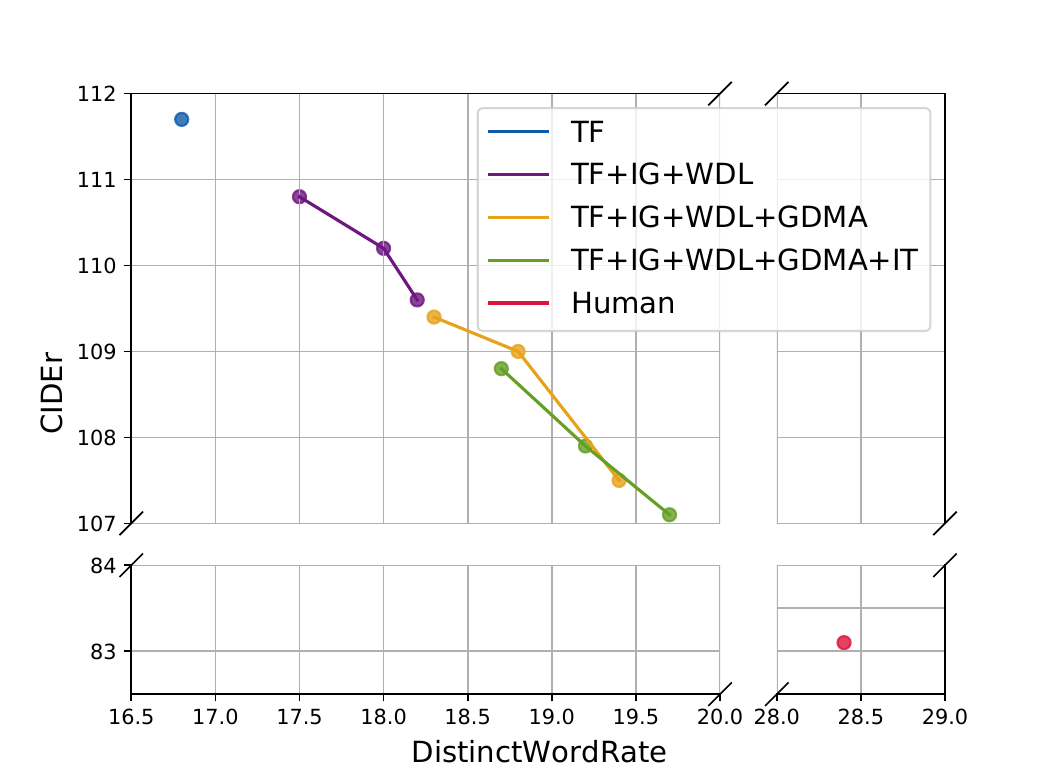}
	\end{center}
	\caption{The trade-off between accuracy (CIDEr) and distinctiveness (DisWordRate): human-annotated GT captions~(Human), baseline  Transformer model (TF)~\citep{2_disccap}, and three variants of our model using various components:
	image group based training (ImageGroup, IG), 
	weighted distinctive loss (WeiDisLoss, WDL), 
	group-based differential memory attention (GDMA), and Indicated Training (IndTrain, IT).
	For our models, we show three training stages (at different epochs), which demonstrates the trade-off between accuracy and distinctiveness during training.
	} 
	\label{fig:trade_off}
\end{figure}

\subsubsection{Comparison with the state-of-the-art}
% \myparagraph{Comparison with the state-of-the-art.} 
We compare our model with two groups of state-of-the-art models: a) FC~\citep{rennie2017self}, Att2in~\citep{rennie2017self}, UpDown~\citep{updown} and AoANet~\citep{huang2019attention} that aim to generate captions with high accuracy; b) DiscCap~\citep{2_disccap}, CL-Cap~\citep{1_contrastive}, and CIDErBtwCap~\citep{17_wang2020compare} that generate distinctive captions. 

The main experiment results are presented in Table~\ref{table:main_results_shrink}, and we make the following observations.
First, when applied to four baseline models, our model achieves impressive improvement for the distinctive metrics, while maintaining comparable results on accuracy metrics. For example, we improve the DisWordRate by $17.3$ percent~(from $16.8\%$ to $19.7\%$) and reduce the CIDErBtw by $5.5$ percent~(from $74.8$ to $70.7$) for Transformer, while only sacrificing the CIDEr by $4.1$ percent. Second, \abcaddpre{in terms of distinctiveness,} models trained with cross-entropy loss tend to perform better than models trained with SCST ~\citep{rennie2017self}. For example, we achieve the highest DisWordRate with Transformer + DifDisCap at $19.7\%$. M$^2$Transformer + DifDisCap also achieves higher DisWordRate than M$^2$Transformer + SCST + DifDisCap.
% \jimmy{Wenjia, please have a look.}\revise{Third, if we use the RegionCLIP feature~\citep{zhong2022regionclip} instead of the Buttom-up feature~\citep{updown}, the models would achieve better performance on both distinctiveness and accuracy since the RegionCLIP is trained with a larger dataset GoogleCC 3M~\citep{sharma2018conceptual}. This leads to the lowest CIDErRank score as 2.20 for Transformer + SCST + DifDisCap + RegionCLIP.}
\revise{Third, compared with state-of-the-art models that improve the accuracy of generated captions, our model M$^2$Transformer + SCST + DifDisCap achieves comparable accuracy, while gaining impressive improvement in distinctiveness. For instance, we obtain comparable CIDEr with AoANet ($127.2$ vs $128.6$), and attain significantly higher DisWordRate, i.e., $14.6\%$ (AoANet) vs. $18.7\%$ (ours).}
When compared to other models that focus on distinctiveness, M$^2$Transformer + SCST + DifDisCap achieves higher distinctiveness by a large margin---we gain DisWordRate by $18.7\%$ compared with $15.9\%$ (CIDErBtwCap), and we obtain much lower CIDErBtw by $80.1$ vs. $89.2$ (DiscCap).

\begin{table*}[h]
\resizebox{\textwidth}{10mm}{
\begin{tabular}{c|ccc|ccc|ccc}
\textbf{Method} & \multicolumn{1}{l}{\textbf{BERTScore}$\uparrow$} & \multicolumn{1}{l}{\textbf{CLIP-S}$\uparrow$} & \multicolumn{1}{l|}{\textbf{PAC-S}$\uparrow$} & {\textbf{DisWordRate} $\uparrow$} & \multicolumn{1}{l}{\textbf{CIDErRank}$\downarrow$} & \multicolumn{1}{l|}{\textbf{CIDErBtw}$\downarrow$} & \multicolumn{1}{l}{\textbf{CIDEr}$\uparrow$} & \multicolumn{1}{l}{\textbf{BLEU3}$\uparrow$} & \multicolumn{1}{l}{\textbf{BLEU4}$\uparrow$}  \\ \hline
Transformer+SCST & \textbf{0.346} & 0.593 & 0.792 & 14.7 & 2.38 & 83.2 & \textbf{127.6} & \textbf{51.3} & \textbf{38.9} \\
+DifDisCap & 0.343 & \textbf{0.595} & \textbf{0.793} & \textbf{17.0} & \textbf{2.34} & \textbf{81.5} & 126.9 & 50.6 & 38.4 \\ \hline
M2Transformer+SCST & \textbf{0.343} & 0.602 & 0.798 & 17.3 & 2.38 & 82.9 & \textbf{128.9} & \textbf{50.6} & \textbf{38.7} \\
+DifDisCap & 0.342 & \textbf{0.603} & \textbf{0.800}  & \textbf{18.7} & \textbf{2.30} & \textbf{80.1} & 127.2 & 49.9 & 38.2
\end{tabular}
}
\caption{\revise{Comparison of model performance on additional metrics, i.e., BERTScore~\citep{zhang2019bertscore}, CLIP-S~\citep{hessel2021clipscore}, and PAC-S~\citep{sarto2023positive}. We experiment on four models (i.e., Transformer+SCST, Transformer+SCST+DifDisCap,  M$^2$Transformer+SCST, and M$^2$Transformer+SCST+DifDisCap).}}
\label{tab:zsl_metric}
\end{table*}

\begin{table*}[h]
\resizebox{\textwidth}{15mm}{
\begin{tabular}{cc|ccc|ccc}
\textbf{Method} & \textbf{$K$} & \multicolumn{1}{l}{\textbf{DisWordRate} $\uparrow$} & \multicolumn{1}{l}{\textbf{CIDErRank}$\downarrow$} & \multicolumn{1}{l|}{\textbf{CIDErBtw}$\downarrow$} & \multicolumn{1}{l}{\textbf{CIDEr}$\uparrow$} & \multicolumn{1}{l}{\textbf{BLEU3}$\uparrow$} & \multicolumn{1}{l}{\textbf{BLEU4}$\uparrow$}  \\ \hline
M$^2$Transformer+SCST & \multicolumn{1}{c|}{-} & 17.3 & 2.38 & 82.9 & \textbf{128.9} & \textbf{50.6} & 38.7 \\ \hline
 & 1 & 17.6 & 2.36 & 81.6 & 128.3 & 50.4 & \textbf{38.8} \\
 & 3 & 18.2 & 2.35 & 81.3 & 127.8 & 50.1 & 38.4 \\
 & 5 & \textbf{18.7} & \textbf{2.30} & 80.1 & 127.2 & 49.9 & 38.2 \\
 & 7 & 18.4 & 2.32 & \textbf{79.9} & 126.9 & 49.8 & 38.1 \\
\multirow{-5}{*}{M$^2$Transformer+SCST+DifDisCap} & 10 & 18.6 & 2.33 & 80.9 & 127.4 & 50.0 & 38.2
\end{tabular}
}
\caption{\revise{The performance on different group size $K$. We experiment on M$^2$Transformer+SCST+DifDisCap model, and also include the performance of M$^2$Transformer+SCST model.}}
\label{tab:K}
\end{table*}

\begin{table*}[h]
\resizebox{\textwidth}{11mm}{
\begin{tabular}{c|ccc|ccc}
\textbf{Method} & \multicolumn{1}{l}{\textbf{DisWordRate} $\uparrow$} & \multicolumn{1}{l}{\textbf{CIDErRank}$\downarrow$} & \multicolumn{1}{l|}{\textbf{CIDErBtw}$\downarrow$} & \multicolumn{1}{l}{\textbf{CIDEr}$\uparrow$} & \multicolumn{1}{l}{\textbf{BLEU3}$\uparrow$} & \multicolumn{1}{l}{\textbf{BLEU4}$\uparrow$}  \\ \hline
Transformer+SCST & 9.4 & 2.47 & 45.7 & \textbf{65.6} & \textbf{34.9} & \textbf{23.6} \\
+DifDisCap & \textbf{11.7} & \textbf{2.28} & \textbf{42.0} & 64.9 & 34.4 & 23.0 \\ \hline
M$^2$Transformer+SCST & 10.9 & 2.52 & 43.4 & \textbf{66.9} & \textbf{35.2} & \textbf{24.1} \\
+DifDisCap & \textbf{12.1} & \textbf{2.17} & \textbf{40.2} & 65.2 & 34.5 & 23.3
\end{tabular}
}
\caption{\revise{The performance of our DifDisCap on Flickr30k dataset. We experiment on four models (i.e., Transformer+SCST, Transformer+SCST+DifDisCap,  M$^2$Transformer+SCST, and M$^2$Transformer+SCST+DifDisCap).}}
\label{tab:flickr}
\end{table*}

\subsubsection{Ablation study}
% \myparagraph{Ablation study.} 
To measure the influence of each component in our DifDisCap, we design an ablation study where we train the baselines M$^2$Transformer and M$^2$Transformer+SCST
\abcadd{with various components of our DifDisCap}.
Four variants of DifDisCap are trained by gradually adding the components, \ie,  ImageGroup (image group based training), WeiDisLoss (weighted distinctive loss), GDMA (group-based differential memory attention), and IndTrain (Indicated Training), to the baseline model. The results are shown in Table~\ref{table:ablation_shrink}, and demonstrate that the four additional components improve the distinctive captioning metrics consistently. As pointed out in~\citet{19_wang2019describing}, increasing the distinctiveness of generated captions sacrifices the accuracy metrics such as CIDEr and BLEU, since the distinctive words cannot agree with all the GT captions due to the diversity of human language. Applying our model on top of M$^2$Transformer increases the DisWordRate by $15$ percent (from $16.4\%$ to $18.8\%$), while only sacrificing $1.8$ percent of the CIDEr value (from $111.8$ to $109.8$). Similarly, applying the proposed modules on M$^2$Transformer+SCST boosts the distinctiveness consistently. Notably, applying the weighted distinctive loss~(WeiDisLoss) improves the DisWordRate from 17.3\% to 18.3\%, which demonstrates the importance of highlighting the distinct words in the GT captions. Moreover, adding the group-based differential memory attention~(GDMA) and Indicated Training~(IndTrain) increases the DisWordRate to 18.7\%, which demonstrates that highlighting the distinct visual information in the target images helps the model to generate distinctive captions.

\revise{We evaluate the influence of image features by replacing the bottom-up features~\citep{updown} extracted by Fast R-CNN~\citep{renNIPS15fasterrcnn} by two alternatives, \ie, ViLBERT~\citep{lu2019vilbert} features and RegionCLIP~\citep{zhong2022regionclip} features. ViLBERT~\citep{lu2019vilbert} learns
task-agnostic joint representations of image content and natural language by extending popular BERT architecture to a multi-modal two-stream model and interacting through co-attentional transformer layers.  RegionCLIP~\citep{zhong2022regionclip} learns region-level visual representations by aligning image regions with template captions in the feature space.}

\revise{As shown in Table~\ref{tab:region_feature}, the RegionCLIP feature achieves superior performance, exemplified by a DisWordRate of 18.9\% and a CIDEr score of 128.2 on the M$^2$Transformer+SCST+DifDisCap model, outperforming other features. This advantage is attributed to its training on a larger dataset, specifically GoogleCC 3M~\citep{sharma2018conceptual}. In contrast, the ViLBERT feature, while offering improved accuracy, results in reduced distinctiveness compared to the Bottom-up feature~\citep{updown}. For instance, on the same model configuration, the CIDEr score marginally increases from 127.2 to 127.8, whereas the DisWordRate decreases from 18.7\% to 18.3\%.}

\revise{We have a view of additional metrics, e.g., BertScore~\citep{zhang2019bertscore}, ViLBERTScore~\citep{lee2020vilbertscore}, V-METEOR~\citep{demirel2022caption}, and UMIC~\citep{lee2021umic}, CLIP-S~\citep{hessel2021clipscore}, and PAC-S~\citep{sarto2023positive}. These metrics could evaluate the semantic similarity between generated captions and target images. We report the performance of BERTScore, CLIP-S, and PAC-S for four models, which is shown in Table~\ref{tab:zsl_metric}. Our DifDisCap leads to higher scores on some no-reference metrics (i.e., CLIP-S and PAC-S) since these metrics measure the similarity between generated captions and target images' visual representation directly. For instance, the CLIP-S score from 0.593 to 0.595 and the PAC-S score from 0.792 to 0.793 by adding DifDisCap to the Transformer+SCST model. Our DifDisCap leads to a slightly worse BERTScore since it still measures the similarity between generated captions and reference captions.}

\revise{Here we also discuss the effect of group size hyper-parameter $K$. As shown in Table~\ref{tab:K}, the results for different $K$ values show that the accuracy and distinctiveness could be trade-offs (e.g., DisWordRate from 17.6\% to 18.7\% and CIDEr score from 128.3 to 127.2 when setting $K$ from 1 to 5). And we use $K=5$ for the best distinctiveness as default. In all, these experiments show that our DifDisCap method is quite robust.}

\revise{In order to evaluate the generalization ability of our model, we further perform experiments on another widely used captioning dataset, i.e., Flickr30k, and display the experiment results in Table~\ref{tab:flickr}. We adapt our DifDisCap on two models (i.e., Transformer+SCST and M$^2$Transformer+SCST). As illustrated in Table~\ref{tab:flickr}, the implementation of the DifDisCap method enhances the distinctiveness of generated captions with minimal impact on accuracy. For instance, upon integrating DifDisCap with the M$^2$Transformer+SCST model, there is a notable increase in DisWordRate (from 10.9\% to 12.1\%) and a slight decrease in CIDEr score (from 66.9 to 65.2) on the Flickr30k dataset. These results mirror the trends observed on the MSCOCO dataset, indicating a consistent performance across different datasets.
}

\subsubsection{Trade-off between accuracy and distinctiveness}
% \myparagraph{Trade-off between accuracy and distinctiveness.}
We next study the accuracy (CIDEr) versus the distinctiveness (DisWordRate) for the models in the ablation study. For comparison, we also compute the CIDEr for the human GT captions, which is the average CIDEr score between one GT caption and the remaining GT captions.
The results in Figure~\ref{fig:trade_off} demonstrate that improving distinctiveness typically hurts the accuracy, since the distinctive words do not appear in all the GT captions, while CIDEr considers the overlap between the generated captions and all GT captions. 
This can explain why the human-annotated GT captions, which are considered the upper bound of all models, only achieve the CIDEr of $83.1$. Compared to the baselines, our work achieves results more similar to human performance.

\section{User Study}
\label{sec:user}

\begin{figure}
	\begin{center}
		\includegraphics[width=\linewidth]{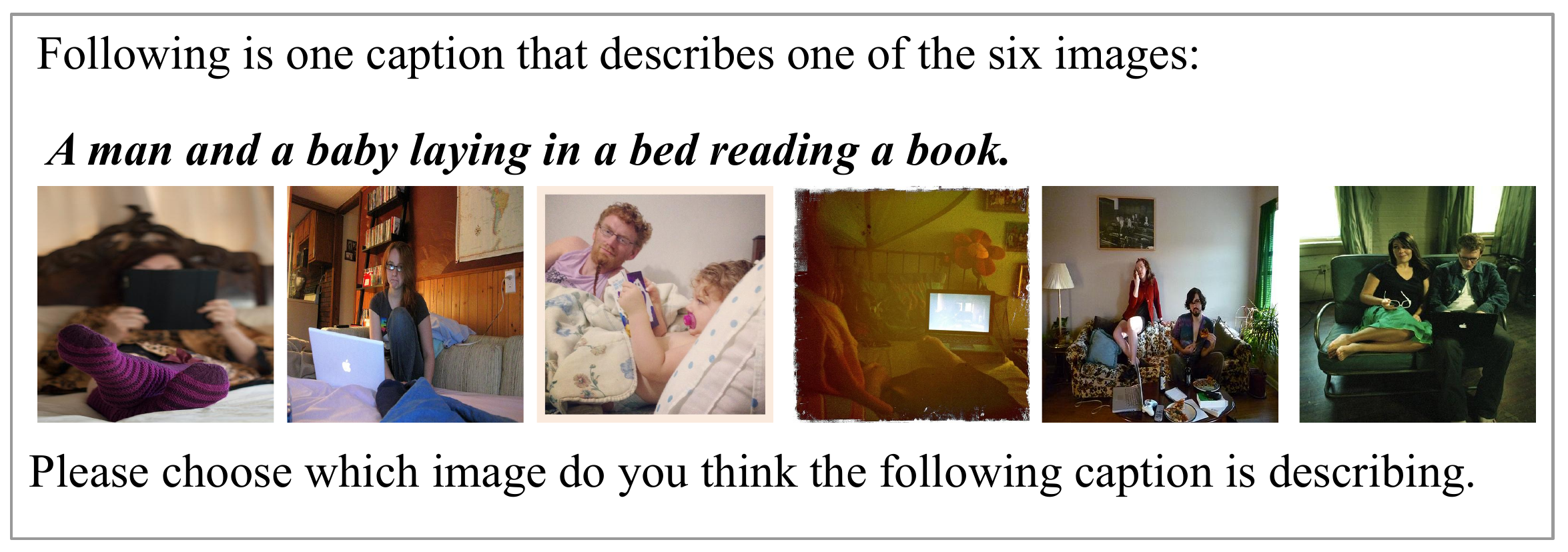}
	\end{center}
	\caption{The user study interface. We display a group of six similar images, and a caption generated from one image by an image captioning model. The users are asked to select the image that they think the caption is describing. }
	\label{fig:user_study}
\end{figure}

To evaluate the distinctiveness of our model from the human perspective, we propose a caption-image retrieval user study, which extends the evaluation protocol proposed in~\citet{17_wang2020compare,2_disccap}. Each test is a tuple $\left( {I_0, I_1, \dots, I_K}, \hat{c} \right)$, which includes a similar image group and a caption generated by a model describing a randomly-selected image in the group. The users are instructed to choose the target image that the caption corresponds to. To evaluate one image captioning model, we randomly select 50 tuples with twenty participants for each test. An answer is regarded as a hit if the user selects the image that was used to generate the caption, and the accuracy scores for twenty participants are averaged to obtain the final retrieval accuracy. A higher retrieval score indicates more distinctiveness of the generated caption, \ie, it can distinguish the target image from other similar images. In Figure~\ref{fig:user_study}, we display the interface for our user study.
\label{section:sp_user_study}

We compare our DifDisCap model with four other models, i.e., DiscCap~\citep{2_disccap}, CIDErBtwCap~\citep{17_wang2020compare},  M$^2$Transformer + SCST~\citep{cornia2020meshed}, and GdisCap~\citep{wang2021group}. The results are shown in Table~\ref{table:user_study}, where our model achieves the highest caption-image retrieval accuracy---$69.5$ compared to M$^2$Transformer+SCST with $61.9$ and GdisCap with $68.2$. The user study demonstrates that our model generates the most distinctive captions that can distinguish the target image from the other images with similar semantics. The results agree with the DisWordRate and the CIDErRank displayed in Table~\ref{table:main_results_shrink}, which indicates that the proposed two metrics are effective evaluations similar to human judgment.

\begin{table}
	\begin{center}
		\small
		\begin{tabular}{c|c}
\textbf{Method} & \textbf{Accuracy} \\
\hline
DiscCap~\citep{2_disccap} & 48.1 \\
CIDErBtwCap~\citep{17_wang2020compare} & 58.7 \\
M$^2$Transformer+SCST~\citep{cornia2020meshed} & 61.9 \\
GdisCap~\citep{wang2021group} & 68.2 \\
DifDisCap  (Ours) & \textbf{69.5}
\end{tabular}
		
		\caption{User study results for caption-image retrieval. 
			Our model produces captions with significantly higher retrieval accuracy (2-sample z-test on proportions, $p<0.01$). 
			%\abc{add citations for the other methods.}
			}
		\label{table:user_study}
	\end{center}
\end{table}

\begin{figure*}[t!]
	\begin{center}
		\includegraphics[width=\linewidth]{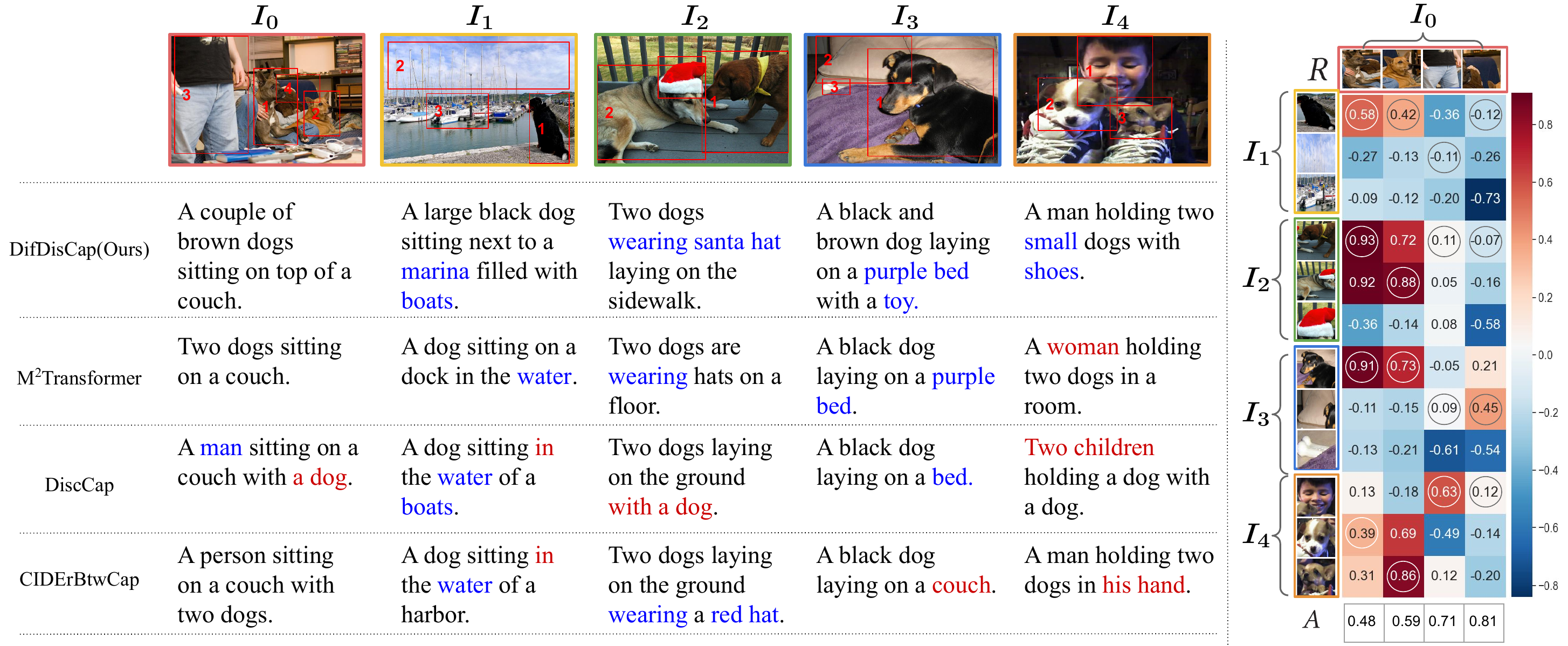}
	\end{center}
	\caption{Qualitative results. Left: Captions for one similar image group with five images from the test set. We compare our model with three state-of-the-art methods, M$^2$Transformer~\citep{cornia2020meshed}, DiscCap~\citep{2_disccap}, and CIDErBtwCap~\citep{17_wang2020compare}. The blue words indicate the distinctive words $\Omega$ that appear in GT captions of the target image, but not in captions of similar images. The red words denote the mistakes in the generated captions. Right: Visualization of the similarity matrix $R$ and distinctive attention $A$. Here $R$ displays the similarity value between four salient objects in $I_0$ and the objects in the other four images, and we use the circle to mark the $\tilde{R}$ value in each group as in (\ref{equ:r_tilde}). The attention $A$ denotes the overall distinctiveness of each object in the image $I_0$. Objects in the same colored box are from the same image.  
	%\abc{it's better to put the $I_0$ image on the far left. Also label the images $I_0$, $I_1$, ... in the (left) and (right) figures.  In the right figure, you could add a circle to mark the $\tilde{R}$ values in the similarity map.} \wenjia{Revised.}
	}
	\label{fig:quali_2}
\end{figure*}

\section{Qualitative Results}
\label{sec:Quali}

In this section, we first qualitatively evaluate our model for generating distinctive captions based on similar image groups. Second, we visualize the group-based memory attention calculated by our model to highlight the distinct objects.
We compare our DifDisCap with three other models, M$^2$Transformer~\citep{cornia2020meshed}, DiscCap~\citep{2_disccap}, and CIDErBtwCap~\citep{17_wang2020compare}, which are the best-competing methods on distinctiveness. More examples are presented in the supplemental.

Figure~\ref{fig:quali_2}~(left) displays the captions generated by the four models for one similar image group. Overall, all the methods generate accurate captions specifying the salient content in the image. However, their performances on the distinctiveness differ.
M$^2$Transformer and DiscCap generate captions that only mention the most salient objects in the image, using less distinctive words. 
For instance, in Figure~\ref{fig:quali_2}~(column 1), our DifDisCap generates captions ``a large black dog sitting next to a marina filled with boats'', compared to the simpler caption  ``a dog sitting on a dock in the water'' from M$^2$Transformer. Similarly, in Figure~\ref{fig:quali_2}~(column 3), our DifDisCap describes the most distinctive property of the target image, the ``santa hat'', compared to DiscCap which only provides  ``two dogs laying on the ground''.
The lack of distinctiveness from M$^2$Transformer and DiscCap is due to the models being supervised by equally weighted GT captions, which tends to produce generic words that agree with all the supervisory captions. 

CIDErBtwCap, on the other hand, reweights the GT captions according to their distinctiveness, and thus generates captions with more distinctive words. Compared to CIDErBtwCap, where all the objects in the image are attached with the same attention, our method yields more distinctive captions that distinguish the target image from others by attaching a higher attention value to the unique details and objects that appear in the image.
For example, in Figure~\ref{fig:quali_2}~(column 3), DifDisCap describes the distinctive ``santa hat'', while  CIDErBtwCap mentions it as a ``red hat''. 

Remarkably, DifDisCap is more aware of the locations of objects in the image and the relationships among them.
For example, in Figure~\ref{fig:quali_2}~(column 5), our caption specifies the ``a man holding two small dogs with shoes'', compared with CIDErBtwCap, which wrongly describes ``holding two dogs in his hand'' when no hands appear on the image. 
It is interesting because there is no location supervision for different objects, but our model learns the relation solely from the GT captions.

Finally, Figure~\ref{fig:quali_2}~(right) displays the similarity matrix $R$ and distinctive attention $A$ for the 2nd image as the target image.  The object regions with the highest attention are those with lower similarity to the objects in other images, in this case the ``couch'' and the ``person''.  The ``dogs'', which are the common objects among the images, have lower non-zero attention so that they are still described in the caption.

\section{Conclusion}
\label{sec:conclude}

In this paper, we have investigated a vital property of image captions---distinctiveness, which mimics the human ability to describe the unique details of images, so that the caption can distinguish the image from other semantically similar images. We presented a Group-based Differential Distinctive Captioning Method (DifDisCap) that compares the objects in the target image to objects in semantically similar images and highlights the unique image regions. 
Moreover, we developed the weighted distinctive loss to train the proposed model. The weighted distinctive loss includes the following two components: the distinctive word loss encourages the model to generate distinguishing information; the memory classification loss helps the weighted memory attention to contain distinct concepts. We conducted extensive experiments and evaluated the proposed model using multiple metrics, showing that the proposed model outperforms its counterparts quantitatively and qualitatively. Finally, our user study verifies that our model indeed generates distinctive captions based on human judgment.

%\abc{any future work?} 
\jimmyadd{\abcaddd{As seen in Figure~\ref{fig:trade_off},} there is still an apparent gap between human and model performance on the distinctive image captioning task. In the future, we will continue leveraging this gap, focusing on both generation and evaluation. Meanwhile, our method could be adapted to other tasks (e.g., distinctive video captioning and contrastive learning in similar image groups).}

\section{Acknowledgments}

This work was supported by a grant from the Research Grants Council of the Hong Kong Special Administrative Region, China (Project No. CityU 11215820).

% use section* for acknowledgment
% \ifCLASSOPTIONcompsoc
  % The Computer Society usually uses the plural form

% Authors must disclose all relationships or interests that 
% could have direct or potential influence or impart bias on 
% the work: 
%
% \section*{Conflict of interest}
%
% The authors declare that they have no conflict of interest.
% \clearpage

% BibTeX users please use one of
\bibliographystyle{spbasic}      % basic style, author-year citations
% \bibliographystyle{spmpsci}      % mathematics and physical sciences
% \bibliographystyle{spphys}       % APS-like style for physics

   % name your BibTeX data base
% \input{refs.bbl}
% Non-BibTeX users please use
% \begin{thebibliography}{}
% %
% % and use \bibitem to create references. Consult the Instructions
% % for authors for reference list style.
% %
% \bibitem{RefJ}
% % Format for Journal Reference
% Author, Article title, Journal, Volume, page numbers (year)
% % Format for books
% \bibitem{RefB}
% Author, Book title, page numbers. Publisher, place (year)
% % etc
% \end{thebibliography}

\end{document}